\documentclass[11pt,a4paper]{article}
\usepackage[hyperref]{emnlp2020}
\usepackage{times}
\usepackage{latexsym}

\usepackage{graphicx}
\usepackage{multirow}
\usepackage{amsmath}
\usepackage{tabularx}
\usepackage{url}
\usepackage{subcaption}
\usepackage{tikz-qtree}
\usepackage{pgfplots}
\usepackage{latexsym}
\newcommand{\argmin}[1]{\underset{#1}{\operatorname{arg}\,\operatorname{min}}\;}
\usepackage{microtype}

\usepackage{microtype}

\aclfinalcopy % Uncomment this line for the final submission
\title{MEGA RST Discourse Treebanks with Structure and Nuclearity \\ from Scalable Distant Sentiment Supervision}

\author{Patrick Huber and Giuseppe Carenini\\
  Department of Computer Science \\
  University of British Columbia \\
  Vancouver, BC, Canada, V6T 1Z4 \\
  {\tt \{huberpat, carenini\}@cs.ubc.ca}}

\date{}

\begin{document}
\maketitle

\begin{abstract}
%One of the major restrictions limiting the effective application of data-driven approaches, such as deep-learning models, for RST-style discourse parsing is the severe lack of large and diverse discourse treebanks annotated with document-level discourse trees. While there are new treebanks being annotated by human linguistic experts, this approach is arguably not scalable to the necessary extend.

%The current 
The lack of large and diverse discourse treebanks hinders the application of data-driven approaches, such as deep-learning, to RST-style
discourse parsing. In this work, we present a novel scalable methodology to automatically generate discourse treebanks using distant supervision from sentiment-annotated datasets, creating and publishing MEGA-DT, a new large-scale discourse-annotated corpus.
%, which we will make publicly available.
%In this paper, we propose an alternative solution to automatically generate large discourse corpora of complete RST-style discourse trees incorporating discourse-structure and -nuclearity.
Our approach generates discourse trees incorporating structure and nuclearity for documents of arbitrary length by relying on an efficient heuristic beam-search strategy, extended with a stochastic component.
%for documents of arbitrary length
%Our approach thereby combines previous work on distant supervision with a novel approach on tree-inference and an effective and efficient heuristic beam search strategy, extended with a stochastic component. 
Experiments on multiple datasets indicate that a discourse parser trained on our MEGA-DT treebank delivers %significantly better 
promising inter-domain performance gains %on the structure and nuclearity measures 
when compared to parsers trained on human-annotated discourse corpora. 
\end{abstract}
%A major limitation for further progress 
%in 

%our novel approach does not only allow the annotation of arbitrary long discourse documents with structure- and nuclearity, \textcolor{red}{but also that  training a discourse parser on the automaticaly annotated large corpus increases its inter-domain performance on both structure and nuclearity prediction.}
\section{Introduction}
\label{Introduction}
%Many NLP tasks nowadays just using deep-neural networks and hope they learn it all\\
%Pre-trained, general purpose encodings like BERT etc. has shown to contain much information on many tasks (probing paper from reading group)\\
%but there is recent evidence that discourse parsing and BERT are complementary (EMNLP paper)\\
%Discourse parsing as an important upstream task \\
%First start with structure, as the most basic task, having the others build on top\\
%Structure already shown to enhance dowstream tasks\\
%However complete discourse trees are better\\
%Existing models are far from being general, due to (probably) overfitting on the small RST-DT dataset\\
%Good performance on that, but performance inter-domain worse.\\
%Mention our previous work \\
%Has been shown to be able to learn automatically \\
%Reaches already better performance on inter-domain discourse structure prediction task \\
%Special importance for nuclearity, as for many downstream tasks, it has been shown that they can be enhanced even without relations \\
%Example approaches were nuclearity chain has been used for summarization \\
%Number of trees is limited to only less than 20. This should be relieved
%Actively selecting attention does make the problem even harder, as more possible trees. \\

Discourse parsing is an important Natural Language Processing (NLP) task, %which aims at uncovering 
aiming to uncover the hidden structure underlying coherent documents, as described by theories of discourse %, typically either 
like Rhetorical Structure Theory (RST) \cite{mann1988rhetorical} or PDTB \cite{prasadpenn}. 
% In this paper, we focus on RST-style pars-ing, but the proposed approach is theory agnostic and could be applied to PDTB as well.
Not only has discourse parsing been shown to enhance %many 
key %critical
downstream tasks, such as text classification \cite{ji2017neural}, summarization \cite{gerani2014abstractive} and sentiment analysis \citep{bhatia2015better,nejat2017exploring,hogenboom2015using}, but it also appears to complement %powerful
contextual embeddings, like 
BERT \citep{devlin2018bert}, in tasks where discourse information
is critical, such as %in 
argumentation analysis \citep{chakrabarty2019ampersand}. 

%RST-style discourse parsing builds a complete, hierarchical discourse tree, traditionally represented as a constituency tree \cite{morey2018dependency}, for a document,

Traditionally, RST-style discourse parsing builds a complete, hierarchical constituency tree for a document \cite{morey2018dependency},
where leaf  nodes  are  clause-like  sentence  fragments, called elementary-discourse-units (EDUs), while  internal  tree nodes  are  labelled  with  discourse relations (e.g., Evidence, Contrast). In addition, each node is given a nuclearity attribute, which encodes the importance of the node in its local context. %While the segmentation of a document into EDUs is a prerequisite for discourse parsing, we do not cover this component here, but use pre-segmented datasets, as further evaluated in section \ref{Evaluation}.

A key limitation for further research in RST-style discourse parsing is the scarcity of training data. Only a few human annotated discourse treebanks exist, each only containing a few %multiple 
hundred documents. 
Although our recent efforts using distant supervision from sentiment to generate large-scale discourse treebanks have already partly addressed this dire situation \citep{huber2019predicting}, the previously proposed solution is %significantly
still limited in: %in multiple aspects: 
%Although recent efforts using distant supervision from sentiment to generate large-scale %discourse-
%treebanks have tried to address this dire situation \citep{huber2019predicting}, the proposed solution is %significantly
%seriously limited in: %in multiple aspects:
(i) Scope, by only building the RST constituency structure without nuclearity and relation labels; and (ii) Applicability, by relying on a non-scalable CKY solution, which cannot be applied to many real-world datasets with especially long documents.

In this work, we propose a significant extension %to the work done by 
to this line of research by introducing a 
%simultaneously boosts scalability, by being applicable to documents of arbitrary length,
scalable solution for documents of arbitrary length
and further 
moving beyond just predicting the tree-structure by incorporating the nuclearity attribute, oftentimes %directly used by
critical in informing
downstream tasks \cite{marcu2000theory, ji2017neural, shiv2019novel}.
%addresses both limitations. We present a novel approach to use distant supervision that moves beyond just the tree-structure of a document, but directly incorporates the key nuclearity attribute, which is oftentimes %directly used by 
%is critical in informing
%downstream tasks (ADD CITATION FOR NUCLEARITY SUMMARY). Our novel approach further allows the computation of documents of arbitrary length.
%and moving beyond just discourse structure, to also predict  nuclearity, and therefore now covering two of the three sub-tasks (structure/nuclearity/relations) of %complete 
%discourse parsing. 
%Our new approach is 
Inspired by the recent success of %applying
heuristic search %approaches
%, like beam search, to %tree-structured %natural language
%to 
in NLP tasks involving trees (e.g., \citet{fried2017improving, mabona2019neural}), %In particular, 
we develop a beam-search strategy implementing an exploration-exploitation trade-off, as
commonly used 
in reinforcement-learning (RL)
\citep{poole2010artificial}. %We intuitively justify this selection of a common RL technique through the similarity of our novel method with RL, both having sparse and noisy %discourse parsing with distant supervision rewards/supervision-signals.

Remarkably,
by following this heuristic approach, any large corpus annotated with sentiment can be turned into a discourse treebank on which a domain/genre specific discourse parser can be trained. As a case study for this process, we annotate, evaluate and publicly release a new discourse-augmented Yelp '13 corpus \citep{tang2015document} called MEGA-DT\footnote{Our new Discourse Treebank and the code to generate further ``silver-standard" discourse treebanks can be found at: \\\url{https://www.cs.ubc.ca/cs-research/lci/research-groups/natural-language-processing/}}  (comprising $\approx$250,000 documents) %, with up to 150 EDUs) 
with nuclearity attributed ``silver-standard" discourse trees, solely leveraging the corpus' document-level sentiment annotation. %With our novel heuristic assumption reducing the computational complexity, the additional nuclearity attribute significantly extends the scope of this approach over a solely structure-based methodology, allowing for more sophisticated tree-aggregation mechanisms on documents of arbitrary length.

To evaluate the quality of our newly proposed MEGA-DT corpus, we conduct %the resulting discourse-annotated corpus 
%is assessed in 
a series of experiments. We train %the established and publicly available state-of-the-art 
the top-performing discourse parser by \citet{wang2017two} on  MEGA-DT %our  generated corpus  
%for jointly predicting discourse structure and nuclearity 
and  compare its performance with the same parser trained on previously proposed treebanks. Specifically, we compare our discourse-annotated dataset against a smaller ``silver-standard" treebank \cite{huber2019predicting} containing around $\approx$100,000 documents with $\leq$20 EDUs and two standard human annotated corpora in the news domain (RST-DT) \cite{carlson2002rst} and in the instructional domain \citep{subba2009effective}.

Results indicate that while training a parser on MEGA-DT does not yet match the performance of training and  testing  on  the  same
%, human-annotated, 
treebank  (intra-domain), it does push the boundaries of what is possible with distant supervision. In most cases, training on MEGA-DT delivers statistically significant improvements 
on the arguably more difficult and useful task of inter-domain discourse prediction,  where  a parser  is  trained  on  one  domain and tested/applied to another one.

Overall, this suggests that our new approach to distant supervision from sentiment can generate large-scale, high-quality treebanks, with MEGA-DT %our novel corpus 
being the best publicly available resource for training a discourse parser in domains where no gold-standard discourse annotation is available.

\section{Related Work}
\label{Related_Work}
%Related Areas: Discourse parsing, Distant supervision from sentiment, heuristic search algorithms (each of this should have roughly equal space) 
%Discourse parsing: describe the key three steps span/structure, nuclearity and relations... current approaches that we will compare with in the evaluation (from EMNLP paper?)... lack of large datasets

%Our work is closely related to three streams of previous research, combining their methodologies, best practices and approaches. 
The most closely related line of work is RST-style discourse parsing, with the goal to obtain a complete discourse tree, including structure, nuclearity and relations. Based on the observation that these three aspects are correlated, most previous work has explored models to learn them jointly (e.g., \citet{joty2015codra, ji2014representation, yu2018transition}). However, while this strategy seems intuitive, the state-of-the-art (SOTA) system on structure-prediction 
%Discourse trees generated in accordance with the RST discourse theory thereby encompasses three components: (1) The tree structure itself, often referred to as the tree span, (2) nuclearity, encoding the local relationship between a internal nodes' subtrees as Nucleus or Satellite, where the nucleus relation implicates the subtree to be of primary importance for the local discourse, while a Satellite assignment implies supplementary subtrees. (3) every internal node contains a relation attribute, describing the relationship between the subtrees of the node (e.g., Contrast, Evidence).\\
%In traditional RST-style discourse parsing task, all three components are required and clearly are correlated.\\
%Previous work done in the area of discourse parsing mostly takes advantage of this correlation between features (span/nuclearity/relation) and computes the discourse tree jointly for all three features \citep{joty2015codra, ji2014representation}. While this approach seems intuitive, the current state-of-the-art system 
%in the area 
by \citet{wang2017two} applies a rather different strategy, first jointly predicting structure and nuclearity and then subsequently predicting relations. The main %intuition 
motivation behind this separation is the large number of possible output classes when predicting these three aspects together.
%combining the three sub-tasks and predicting them together. %Separating the computation shows improvements, despite the strong intuition of the correlation between features. 
The success of the system by \citet{wang2017two} on the widely used RST-DT corpus inspires us to also learn structure and nuclearity jointly%for the difficult task of discourse parsing 
%from distant supervision
, rather than %focusing on only structure (as done in \citet{huber2019predicting}) or 
combining all three aspects. %sub-tasks.

%In this paper

%following the intuition of previous approaches and limiting the system to sentence-level prediction. Independent of the what intuition is used on how to separate the difficult task of RST-style discourse parsing, the vast majority of previous models are nearly exclusively trained and tested on the RST-DT dataset \citep{mann1988rhetorical}, which only contains a 385 document subset of the Wall-Street Journal corpus. Other corpora, even smaller in size than RST-DT, are the Instructional Corpus by \citet{subba2009effective} and the GUM corpus \citep{Zeldes2017}. The strict size limitation of the available datasets clearly calls for a new, large scale corpus to be used in the area of discourse parsing, to allow deep-learning approaches to be applied in the domain more effectively and to avoid methodologies to overfit on the dataset through elaborate feature engineering.\\

The second line of related work infers fine-grained information from coarse-grained supervision signals using machine learning. Due to the lack of annotated data in many domains and for many real-world tasks, methods to automatically generate reliable, fine-grained data-labels have been explored for many years. One promising approach in this area is %uses the 
Multiple Instance Learning (MIL) %) %framework, 
%initially studied by 
\citep{keeler1991integrated}.
%over two decades ago. 
The general task of MIL is to retrieve fine-grained information (called instance-labels) from high-level supervision (called bag-labels), using correlations of discriminative features within and between bags to predict labels for instances. With the recent rise %upcoming
of deep-learning,
%and the resulting opportunity to learn complex projection functions purely from data without manual feature engineering, 
neural MIL approaches have also been %explored
proposed \citep{angelidis2018multiple}.
%EMNLP paper needs to be summarized here first and then we say how we improve on it....

We previously combined the two lines of related work described above to create discourse structures from large-scale %auxiliary 
datasets, solely using document-level supervision \cite{huber2019predicting}. 
%As such, the authors repeatedly stress the potential of this approach to be applied to many existing, large-scale datasets. 
When applied to the auxiliary task of sentiment analysis, 
%the authors create a large scale 
we generated a ``silver-standard" discourse structure treebank using the neural MIL model by \citet{angelidis2018multiple} %for distant supervision 
in combination with a sentiment-guided CKY-style tree-construction algorithm, generating near-optimal discourse trees in bottom-up fashion \cite{jurafsky2014speech}. 
Although our approach has shown clear benefits, it is inapplicable to many real-world datasets, as it does not scale to long documents and cannot predict nuclearity- and relation-labels.
%been shown to produce useful data to train discourse parsers, it does not scale to %is seriously limited in: (1) Scope, by only building the discourse tree structure without nuclearity and relation labels and (2) Applicability, by relying on the non-scaleable CKY solution, %which cannot be applied  
%inapplicable to many real-world datasets. %problems.
%However, the authors do only consider a single tree property (structure) and approach the problem at hand in a highly non-scaleable way, using exhaustive search, preventing their work from being applied to many real-world problems.
Addressing these limitations is a major motivation of this paper.

Further efforts to automatically generate discourse trees from auxiliary tasks have been mostly
focused on latent tree induction, %as proposed by, 
generating trees from text classification \cite{karimi2019learning} or summarization tasks  \cite{liu2019single}. %, utilizing supervision from summarization. 
For both approaches, domain dependent discourse trees are induced during the neural training process. While either method has shown to improve the performance on the downstream task itself, subsequent research by \citet{ferracane2019evaluating} indicates % the strict limitations of the 
that the induced trees %, which 
are often trivial and shallow, and %therefore unsuitable to train a discourse parser.
do not represent valid discourse structures.

The third stream of related work 
%has successfully explored the use of
is on leveraging
heuristic search algorithms in NLP tasks involving trees. %on tree-structured natural language tasks 
%such as syntactic %constituency parsing. %\citep{fried2017improving} 
%and discourse parsing. %citep{mabona2019neural}.
%enabling  researchers in diverse sub-fields of NLP to efficiently compute relevant parts of the huge space of possible solutions. \\
For syntactic parsing, \citet{vinyals2015grammar} and \citet{fried2017improving} show that a static, small beam size (e.g. 10) %can
already achieves good performance, with \citet{dyer2016recurrent} delivering promising results by using greedy decoding. %(beam size =1). %Similarly, \citet{vinyals2015grammar} show that a beam size of less than 10 is sufficient, and \citet{dyer2016recurrent} even just use greedy decoding for their %discriminative parsing model.
%While the primary focus of  \citet{fried2017improving} is evaluating the disentangling of model combinations and re-ranking effects, they also show that a static, small beam size can already achieve good performance on the task of syntactic constituency parsing. Further work investigating the effects of beam search on syntactic parsing is conducted by \citet{vinyals2015grammar} and \citet{dyer2016recurrent}. \citet{vinyals2015grammar} show that a beam size of less than 10 can already achieve good results, while \citet{dyer2016recurrent} even uses greedy decoding for their discriminative model.
As a recent example %within the area of 
for discourse parsing, \citet{mabona2019neural} successfully combine standard beam-search with shift-reduce parsing  %framework%use an adaptation of the standard beam search algorithm, in combination with the shift-reduce framework, 
%to generate more balanced trees, 
using two parallel beams for shift and reduce actions.
%, removing the left-branch bias. 
%While we are not using the shift-reduce framework in our work, t
Overall, %This recent stream of research 
recent work
%clearly 
shows that %the use of 
beam-search approaches %as such, 
and their possible extensions can effectively address scalability issues in multiple parsing scenarios. In this paper, we %will 
extend the standard 
beam-search approach %methodology 
with a stochastic exploration-exploitation trade-off, as %commonly 
used in Reinforcement Learning, where signals also tend to be sparse and noisy.
%The rationale is that both RL and distant supervision rely on signals that tend to be sparse and noisy.
%The use of this intuition is especially useful, as both tasks (RL and discourse parsing with distant supervision) are facing the problem of sparse rewards, leading to similar problems. We further experiment with the beam size and the heuristic function to use in this framework. 

\section{Predicting Discourse from Sentiment}
% Phrase this stronger, less as an extension more like individual work using this intuition

Previous work has shown that incorporating RST-style discourse trees can help to predict document-level sentiment \cite{ bhatia2015better,nejat2017exploring,hogenboom2015using}. 
These findings give rise to the assumption that the sentiment of a document can also provide important information on its discourse structure. In the following sub-section, we shortly revisit our previous approach to %describe a way to 
exploit this assumption by solely relying on document-level sentiment annotations %, as initially explored in 
\cite{huber2019predicting}. Afterwards, we will present our new approach to overcome scalability issues and jointly predict structure and nuclearity.
%the strict limitations and reasonably extend the approach to jointly predict structure and nuclearity in section \ref{arbitrary_doc}
%In this section we will first describe a small-scale example of the straight-forward approach to create discourse trees with distant supervision. The basic methodology is thereby used as intuition for our work, in which we show the potentials and limitations of the simple approach and solve the pressing problem on how to create discourse trees suitable for real-world data. 

\subsection{Predicting Discourse Structure for Short Documents}
%based on a simple (four EDUs?) example (maybe even including the actual text)... saying something like." In order to predict the discourse structure of a document given its global sentiment, we apply .... get sentiment and attention for each EDUs. For instance for the following document comprising four EDUs we would get... This info is then bla bla CKY... this process return the trees shown in Fig. with the associated attention and sentiment scores on each node... tree x is selected because.... Although this method has been shown to provide reasonably good trees it is limited bla bla 

%The general approach taken 
The discourse-structure tree of a document can be predicted from its global sentiment
%, already described by \citet{huber2019predicting}, 
by combining Multiple Instance Learning (MIL) %\citep{angelidis2018multiple} 
and the CKY algorithm. %\citep{jurafsky2014speech}
We will illustrate the process on the following negative sentiment example with a polarity of $-0.5$, pre-segmented into five EDUs: \textit{[Bad poutine.]\textsubscript{1}, [Fries were nice and thin,]\textsubscript{2}, [had a good taste,]\textsubscript{3}, [however, the gravy was cold]\textsubscript{4}, [and the cheese was not melted.]\textsubscript{5}}. 
The first step in generating the discourse structure for this example consist of assigning each EDU a sentiment polarity $p_{EDU}$ within the interval of $[-1, 1]$ and an attention score $a_{EDU}$ between $[0, 1]$, both learned through MIL from the overall document sentiment polarity. 

To obtain the tuple $\{p_{EDU}, a_{EDU}\}$ for each EDU in a document, the neural MIL model \cite{angelidis2018multiple} is trained on a document-level sentiment dataset, with the goal to predict sentiment-labels for EDU-level instances. The model therefore generates a mapping from inputs (EDUs) to the respective outputs (sentiment-classes) by exploiting correlations between the appearance of EDUs in documents and the respective document gold-labels across a corpus. For example, the EDU \textit{[had a good taste,]\textsubscript{3}} will most likely appear predominantly in positive documents, allowing the MIL model to infer a positive EDU-level sentiment polarity $p_{EDU}$ for this input. When applying the neural model by \citet{angelidis2018multiple}, an attention mechanism is internally used to weight the importance of EDUs for the overall document sentiment. An attention-weight $a_{EDU}$ is also extracted for each EDU and subsequently used as an importance score when aggregating subtrees using the CKY approach.

%The approach predicts the sentiment labels $\hat{Y} = (\hat{y_1}, ..., \hat{y_n})$ of EDU-level instances $X = (x_1, ..., x_n)$ from the gold-standard sentiment scores $Y = (y_1, ..., y_m)$ on document-level (in MIL called bags $B$). Thereby, each bag $b_i \in B \subset \{(X' \subseteq X), y_i\}$ contains a subset of instances $X'$ and has a single sentiment label $y_i$. Depending on the occurrences of $x_j \in X$ in $B' = B(x_j) \subseteq B$, a correlation based on the sentiment labels $Y$ can be inferred, mapping the labels of instances as $m: x_j \in X \rightarrow \hat{y_j}$, with $\hat{y_j}$ referring to the EDU level polarity-score $p_{EDU}$. When using the neural model by \citet{angelidis2018multiple}, an attention mechanism is internally used to weight the importance of EDUs for the overall document sentiment. The attention-weights $a_{EDU}$ are also extracted per EDU and subsequently used as an importance score when aggregating subtrees using the CKY approach.

From those tuples $\{p_{EDU}, a_{EDU}\}$ assigned to leaf-nodes, the sentiment polarity $p$ and attention score $a$ for any internal node in an arbitrary constituency tree can be computed bottom-up by aggregating its two child nodes $c_l, c_r$. Out of the set of potential aggregation functions proposed in \citet{huber2019predicting}, the best performing approach has shown to be:
\begin{equation}
p = \frac{ p_{c_{l}}*a_{c_{l}}+p_{c_{r}}*a_{c_{r}}}{a_{c_{l}}+a_{c_{r}}}\quad a = \frac{a_{c_{l}}+a_{c_{r}}}{2}
\label{eq:functions}
\end{equation}
By recursively applying this function
%, starting 
from the leaf-nodes, we can compute the sentiment and attention of the root node, representing the full document.

\begin{figure}
\setlength{\belowcaptionskip}{-10pt}
    \centering
    \includegraphics[width=1\linewidth]{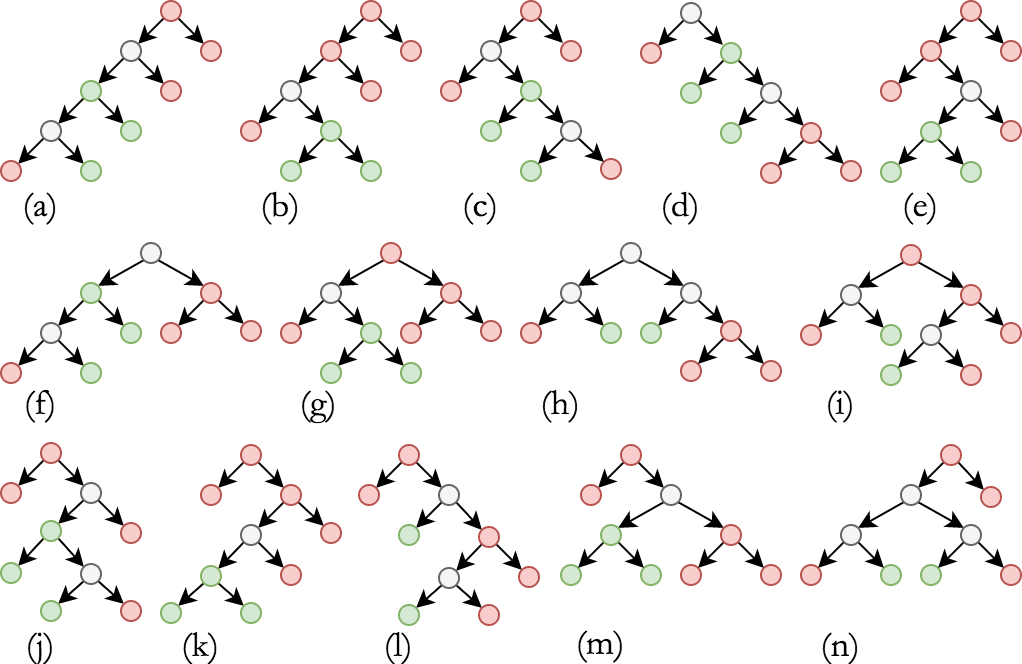}
    \caption{All 14 projective discourse trees annotated with sentiment for a 5 EDU document  (using a simplified color-scheme,  green = positive, red = negative, grey = neutral, omitting the attention attribute)}
    \label{fig:all_permutations}
\end{figure}

The process of selecting the best discourse tree for a given document can be framed as finding the tree for which the sentiment of the root node (spanning the whole document) is the closest to the  gold-standard sentiment annotation.
%for that document. 
A brute-force solution to this problem is to generate all possible
%$\frac{1}{n}\binom{2n-2}{n-1}$ 
%projective 
discourse trees using the general CKY algorithm and selecting the best tree amongst all candidates. However, the computational complexity of this approach quickly explodes, 
as shown for our running example with 5 EDU leaf-nodes in Figure \ref{fig:all_permutations}. From the decision-space of possible tree-structures, the tree 
%shown in Figure \ref{fig:tree} would be selected, because its 
with the shortest sentiment-distance from the gold-standard, computed at the root node, is selected. %($(g)$ in this example).
%In this example (using a simplified color-scheme) the optimal tree to select is $(g)$.

%the in  A possible discourse tree chosen based on distant supervision from the gold-label sentiment 
%and the leaf-node sentiment-attention tuples $\{p_{EDU}, a_{EDU}\}$ 
%is illustrated in Figure \ref{fig:tree}.
%selecting

%The most straight-forward way to find a discourse tree closely resembling the original authors sentiment is by trying all possible $\frac{1}{n}\binom{2n-2)}{n-1}$ projective trees using a dynamic programming approach, such as CKY \citep{jurafsky2014speech}. An example decision space using this approach for a discourse tree with 5 leaf-nodes is shown in Figure \ref{fig:all_permutations}. A possible discourse tree chosen based on distant supervision from the gold-label sentiment 
%and the leaf-node sentiment-attention tuples $\{p_{EDU}, a_{EDU}\}$ 
%is illustrated in Figure \ref{fig:tree}.

%\begin{figure}
%    \centering
%    \includegraphics[width=0.99\linewidth]{single_tree.png}
%    \caption{Possible discourse tree based on distant supervision from gold-label sentiment (center second row in Figure \ref{fig:all_permutations})}
%    \label{fig:tree}
%\end{figure}

%The tree shown in Figure \ref{fig:tree} is selected due to its close resemblance of the authors (negative) sentiment for the document. While all $\frac{1}{n}\binom{2n-2)}{n-1}$ trees have been computed with the CKY algorithm, all other possible trees have been determined inferior to the one shown, due to the heuristic definition of a discourse tree being able to resemble the document's original sentiment. 

Although this method has been shown to provide reasonably good trees %for short documents 
when leveraged for discourse-structure parsing, it is limited in two fundamental ways: %\\ %aspects. 
(1) The approach is not %well 
scalable. Since its space complexity grows with %is quantified by 
the Catalan number $C_n = \frac{1}{n+1}\binom{2n}{n}$ for trees with $n+1$ EDUs, it can %consequently 
only be applied to short ($\approx \leq$ 20 EDUs) documents (see bottom row in Table \ref{tab:beam_upper_bound}),
making it impractical for many %interesting 
real-world datasets %which contain large numbers of
containing longer documents, such as the Yelp '13 \citep{tang2015document}, IMDB \citep{diao2014jointly} or Amazon Review dataset \citep{zhang2015character}.
%used for small subsets of arbitrary datasets containing very short documents (up to $\approx20$ EDUs per document). 
%This prohibits the general methodology to be used for many %interesting 
%real-world datasets annotated with sentiment, %applications
%containing far beyond the feasible number of EDUs in a single document, such as the Yelp '13 \citep{tang2015document}, IMDB \citep{diao2014jointly} or Amazon Review \citep{zhang2015character} datasets.\\
%summarization datasets (CNN/DailyMail \citep{RNN_beyond}, PubMed \citep{cohan-etal-2018-discourse}, ArXiv \citep{cohan-etal-2018-discourse}), question-answering corpora like SQuAD \citep{rajpurkar2016squad, rajpurkar2018know}, aspect-based sentiment analysis datasets \citep{tay2018learning} and long sentiment datasets, such as the complete Yelp '13 corpus \citep{tang2015document}, IMDB \citep{diao2014jointly} or Amazon Reviews \citep{zhang2015character}.\\
%\textbf{(2)} Already for short documents, the computational efforts increase according to the Catalan number, defined as $C_n = \frac{1}{n+1}\binom{2n}{n}$, which makes the approach highly non-scaleable.\\
%(3) nuclearity? .... \\
(2) Due to the high computational complexity of the structure prediction itself, the inference of further RST-tree properties, such as nuclearity and relations, often critical for downstream tasks, are not feasible with this unconstrained CKY approach.

\subsection{Predicting Discourse Structure and Nuclearity from Arbitrary Documents}
\label{arbitrary_doc}
%Add the theoretical upper-bounds for CKY and 
Inspired by the recent success in applying beam-search to enhance the scalability of multiple NLP parsing tasks \citep{mabona2019neural, fried2017improving, dyer2016recurrent, vinyals2015grammar}, we propose a novel heuristic beam-search approach %in order to
that can automatically generate discourse trees containing structure- and nuclearity-attributes for documents of arbitrary length.

\paragraph{Stochastic Beam-Search}
%In essence, the CKY dynamic-programming algorithm used to create all the binary trees covering $n$ EDUs, does that by filling an $n.n$ matrix, where each $cell(i,j)$ contains all the subtrees covering the text span from $EDU_i$ to $EDU_j$. Our heuristic beam search solution limits the computational complexity of this process by fixing the number of subtrees that can be stored in each cell to a constant beam-size.

In essence, the general CKY dynamic programming algorithm creates all possible binary trees covering the $n$ EDUs by internally filling an $(n \times n)$ matrix, where each $cell(i,j)$ contains the information on all subtrees covering the text spans from $EDU_i$ to $EDU_j$. Our heuristic beam-search solution 
%strategy 
limits the computational complexity of this process by reducing the number of subtrees stored in each cell to a constant beam-size $B$.
%, in the spirit of recent successes in using beam search to heuristically reduce the number of necessary computations
%To evaluate the beam search approach for the problem at hand we explore two versions of the heuristic beam search approach:
%Limiting the number of subtrees that are kept in each cell of the CKY dynamic programming matrix.
%Intuition
This naturally raises the question on how to select the $B$ subtrees to preserve in each cell. We follow the intuitive assumption that subtrees for which the sentiment diverges most from the overall document sentiment (the only available supervision for this task) can be safely discarded. 
%This suggests an initially rather strict rule for subtrees to be included in the beam. 
Out of the set of possible subtrees $T$ for a given cell, only the subset $T'$ with $|T'| = B$ is kept, containing the $B$ subtrees with the closest sentiment polarity $p_{t_i}$ to the gold-label ($gl$) sentiment of the document. Formally: %(see equation \ref{eq:beam_basic} for the formal definition of the beam-selection).
%(see eq. \ref{eq:beam_basic}).
\begin{equation}
    T' = \argmin{t_i \in T, |T'| = B} |p_{t_i}-gl|
\label{eq:beam_basic}
\end{equation}

However, one limitation of this heuristic rule %presented above 
is that it strictly prefers subtrees with sentiment closer to the overall document sentiment, independent of their distance from the root node. This can be problematic when applied in early stages of the tree-generation process, where only a few EDUs are combined. For instance, a mostly positive document might still contain certain negative subtrees at its lowest levels, which also need to be aggregated appropriately.
%the constant reduction rate strictly based on the gold label sentiment, not taking the current level and therefore the distance of the current node to the root-node into account. One can easily convince themselves that this is problematic when applied at early stages of the tree-generation process, where only a few EDUs are combined. 

Ideally, we would like to support a high degree of exploration on low levels of the tree, only loosely forcing the sentiment of subtrees in the beam to align with the overall document gold-label sentiment; while on higher levels of the tree, the requirement of closely reflecting the document's gold-standard sentiment should be strictly enforced (i.e., exploiting the distant supervision).

%we exploit the computed distance between subtrees and the gold-label sentiment, strictly enforcing resulting beams to closely represent the document's gold-standard sentiment and only allowing a low degree of exploration into unlikely trees.
%To avoid enforcing excessively  strict limitations to subtrees on low levels of the overall tree, we further explore stochastically optimized beams, introducing a dynamic exploration-exploitation trade-off \citep{poole2010artificial} based on the relative level of the node in the overall tree structure. The goal is 
%to allow a high degree of exploration on low levels of the tree, only loosely forcing the beam to align with the overall document gold-label sentiment, while on higher levels of the tree, we exploit the computed distance between subtrees and the gold-label sentiment, strictly enforcing resulting beams to closely represent the document's gold-standard sentiment and only allowing a low degree of exploration into unlikely trees. 

We implement this strategy through a stochastic beam-search approach, which relies on a softmax selection using the Boltzmann–Gibbs distribution \citep{poole2010artificial}. The temperature coefficient $\tau$ thereby modulates %regularize 
the exploration-exploitation trade-off (similar to previous work in RL), by influencing the divergence of the softmax outputs.
We then sample from the resulting, categorical probability distribution $P = ((Prob(t_1), ..., Prob(t_n))$, computed for every local subtree $t_i \in T$ to obtain a subset $T'$ of size $B$ (as shown in equation \ref{eq:tradeoff}).

\begin{figure}
\setlength{\belowcaptionskip}{-10pt}
    \centering
    \includegraphics[width=\linewidth]{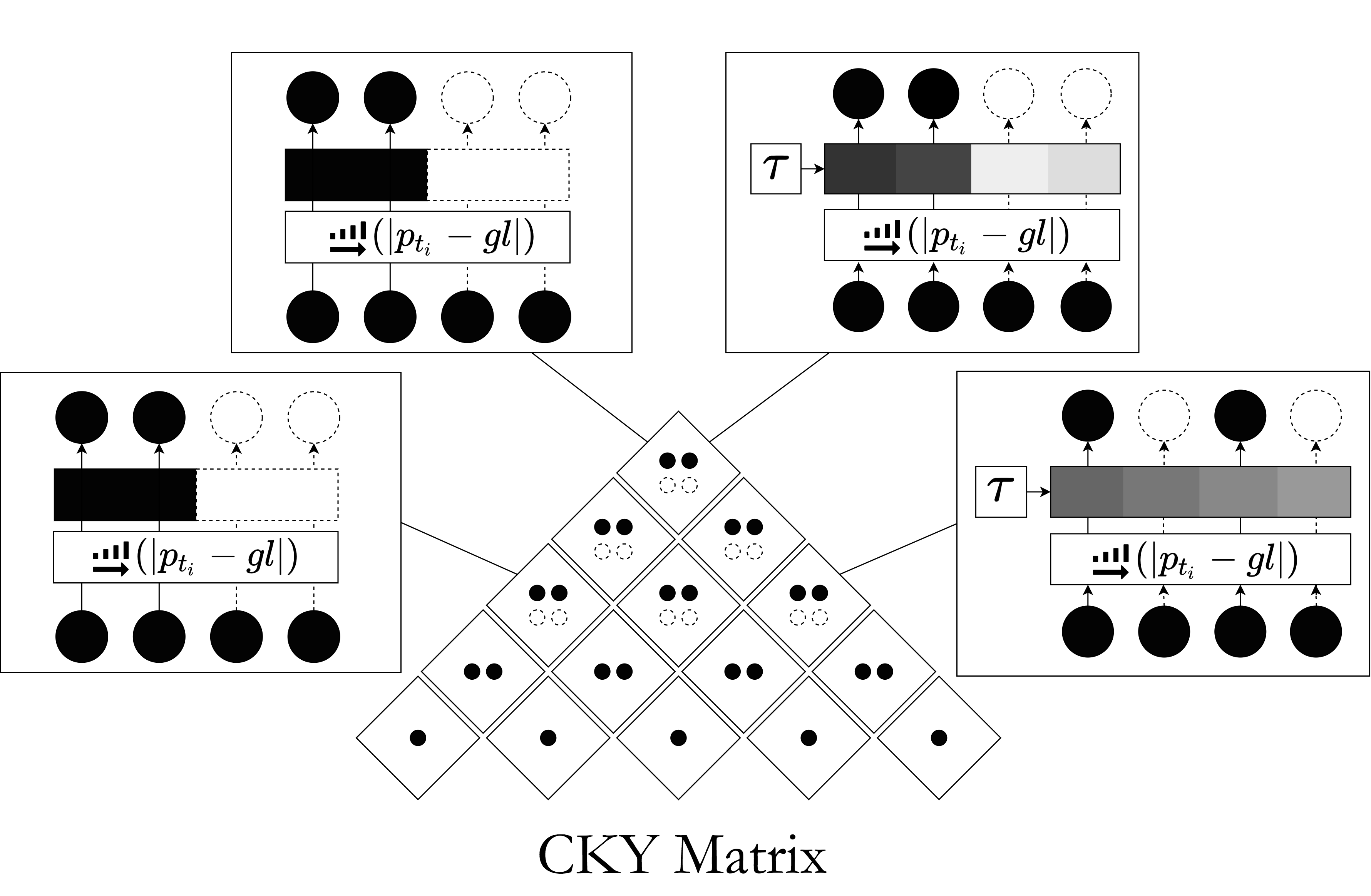}
    \caption{Standard beam-search approach (left) picking the top $B=2$ tree-candidates with the smallest distance $|p_{t_i}-gl|$ in every CKY cell. Stochastic beam-search approach (right) calculating the Boltzmann–Gibbs distribution with the tree-coverage dependent temperature $\tau$, modulating the subtree sampling process of the tree-candidates. (For readability, we only show a maximum of 4 subtrees per CKY cell)}
    \label{fig:std_tree}
\end{figure}

\begin{equation}
    Prob(t_i) = \frac{e^{\frac{1}{|p_{t_i}-gl|}/\tau}}{\sum\limits_{t_j \in T}e^{\frac{1}{|p_{t_j}-gl|}/\tau}}
\label{eq:tradeoff}
\end{equation}
\begin{equation}
%\tau = f(h,m) = \frac{1}{|m-h|+1} 
\tau = f(n,c) = (n-c)+1
\label{eq:tau}
\end{equation}

In this work, the parameter $\tau$
%, defining the exploration-exploitation trade-off 
is defined as a linear function $f(n,c)$ parameterized by the number of EDUs $c$ covered under the subtree $t_i$
%rooted in the current node $t_i$ 
as well as the total number of EDUs $n$ (see equation \ref{eq:tau}). This way, $\tau$ influences equation \ref{eq:tradeoff} such that for larger values of $c$ (at the top of the tree), $\tau$ gets close to 1 and the sampling is likely to select subtrees with low distance $|p_{t_i}-gl|$. For subtrees with a small coverage $c$ (at the bottom of the tree), $\tau$ becomes $>>1$ and $Prob(t_i)$ resembles the uniform distribution, allowing for a high degree of exploration.
%As already mention in section \ref{Related_Work}, a comparable approach of the trade-off between exploration and exploitation is often used in the area of reinforcement-learning, where rewards are sparse (just like they are in our distant-supervision scenario) to enable the agent to find a good solution. 
For illustration, Figure \ref{fig:std_tree} 
%and \ref{fig:stoch_tree} 
highlights the differences between the standard and the stochastic beam-search approach. %respectively.

%\begin{figure}
%    \centering
%    \includegraphics[width=0.8\linewidth]{procedure_std.png}
%    \caption{Standard beam-search approach picking the top $B=2$ tree-candidates with the smallest distance $|p_{t_j}-gl|$ at every level of the tree}
%    \label{fig:std_tree}
%\end{figure}

%\begin{figure}
%    \centering
%    \includegraphics[width=0.99\linewidth]{precedure_stoch.png}
%    \caption{Stochastic beam-search approach using a  softmax-function based on the Boltzmann–Gibbs distribution ($G/B$), a tree-level parameter $\tau$ and a uniform distribution ($U$). At every level of the tree, a stochastically weighted sample is picked based on the exploration-exploitation trade-off (see selection-gradient  visually shown in each node)}
%    \label{fig:stoch_tree}
%\end{figure}

\paragraph{Analysis of Spacial Complexity}
The described system significantly reduces the spatial complexity, independent from whether a stochastic component is used. The complexity reduction can be easily observed by comparing the theoretical upper-bounds for the space consumption of the unrestricted CKY approach (eq. \ref{eq:cky}) against the upper-bounds for the heuristically constrained CKY method (eq. \ref{eq:beam}).

\begin{equation}
    \sum_{i = 1}^{n-1}{\frac{4(n-i)}{i}\binom{2i-2}{i-1}}
\label{eq:cky}
\end{equation}
\begin{equation}
    4n^2B + 4(n-1)B^2
\label{eq:beam}
\end{equation}

%∑(4*(100-i))/i * ((2*i-2) binom (i-1)), i=1 to 99 Wolfram
In both equations, $n$ represents the number of leaf-nodes (EDUs) in the discourse tree. In eq. \ref{eq:cky}, the number of generated trees at every level of the tree is bound by the Catalan number, while in eq. \ref{eq:beam} the bound has a quadratic dependency on the input-size and the beam-size.
For the equations shown, we assume that on every level of the tree, each of the possible subtrees is represented by 2 pointers to the child-nodes as well as a sentiment and attention value for the subtree itself.
Table \ref{tab:beam_upper_bound} compares the space capacities required with increasing document length, indicating that with a proper beam size, 
our heuristic strategy can deal with the tree structures  for very long documents. 
%Unconstrained CKY (i.e., Beam size equal to $\infty$) is compared with heuristic solutions with increasing beam sizes.

%shows the space capacity required regarding the beam size $B$ and the document length in EDUs, also compared to the unconstrained CKY approach.

%On every level of the tree, each of the possible subtrees is represented by 4 pieces of information: 2 pointers to the child-nodes of the subtree as well as a sentiment and attention value for the subtree itself. Assuming that every information piece takes a single unit of memory, the amount of information stored for a discourse tree with 5 EDUs is $64$. For a larger discourse tree of 20 EDUs $3.6 * 10^9$ units of information need to be stored in memory, for 30 EDUs the required memory capacity is $1.9 * 10^{15}$ and for a document with 100 EDUs, the full CKY approach stores $4 * 10^{56}$ pieces of information, far beyond what is possible with modern computers.\\
%As a comparison, using beam-search with a beam size of $B$, the search space is greatly reduced, with only a polynomial dependency between the input length (number of EDUs) and the amount of information stored in memory. Table \ref{tab:beam_upper_bound} shows the space capacity required regarding the beam size $B$ and the document length in EDUs, also compared to the unconstrained CKY approach. 
%Please note that the upper bound given in equation \ref{eq:beam} is not tight for small examples, which, in those cases, are also upper-bounded by equation \ref{eq:cky}. For the ease of understanding, we still show them in table \ref{tab:beam_upper_bound}.

\begin{table}
\setlength{\belowcaptionskip}{-10pt}
    \centering
    \begin{tabular}{|r|r r r|}
        \hline
        Beam & 20 EDUs & 30 EDUs & 100 EDUs \\
        \hline\hline
        %$1$ & $1,676$ & $3,716$ & $40,396$ \\
        %$10$ & $2.4*10^4$ & $4.8*10^4$ & $4.4*10^5$ \\
        %$100$ & $9.2*10^5$ & $1.5*10^6$ & $7.9 * 10^6$ \\
        %\infty & $3.6*10^9$ & $1.9*10^{15}$ & $4 * 10^{56}$ \\
        $1$ & $1.6$KB & $3.7$KB & $40$KB \\
        $10$ & $24$KB & $48$KB & $440$KB \\
        $100$ & $920$KB & $1.5$MB & $7.9$MB \\
        $\infty$ & $3.6$GB & $1.9$PB & $400$SB \\
        \hline
    \end{tabular}
    \caption{Upper-bounds for growth of spatial complexity using different beam sizes and unconstrained CKY ($\infty$), assuming 1Byte per unit in memory. KB = $10^3$, MB = $10^6$, GB = $10^9$, PB = $10^{15}$, SB = $10^{54}$}
    \label{tab:beam_upper_bound}
\end{table}

%Recalling that memory is the main restriction of the complete CKY approach employed by \citet{huber2019predicting}, as the CKY approach itself already reduces the time complexity of the computation to a polynomial dependency of $n^3$ in the input through dynamic programming, the addition of a heuristic search component greatly improves the memory consumption and allows for an efficient computation of large-scale documents.\\

\paragraph{Integration of Nuclearity}
With this %newly implemented
scalable solution, it is now possible to also take additional properties, like nuclearity, into account.
%even if this introduces further computational complexity. 
The inherent advantage of generating nuclearity-attributed discourse trees becomes obvious when revisiting the definition %of nuclearity 
in RST \cite{mann1988rhetorical}, where the nuclearity-attribute encodes a notion of ``importance" in the local context, with \textit{Nucleus-Statellite (N-S)} and \textit{Satellite-Nucleus (S-N)} attributions defining the directionality between two nodes, while the \textit{Nucleus-Nucleus (N-N)} attribution implies equal importance \cite{morey2018dependency}. Expressing this notion of importance, it is not surprising that  nuclearity-attribution is frequently critical  in  informing  many downstream tasks like summarization and text categorization  (e.g., \citet{marcu2000theory, ji2017neural, shiv2019novel}).

%Our newly implemented scalable solution 
%opens up the possibility of taking additional properties of the tree into account. This has previously been intractable when using the unconstrained CKY approach, as any additional attribute (nuclearity/relations) introduces further computational complexity. In this paper, we are extending the plain, structure-based discourse trees with an additional nuclearity attribute to refine the purely structure-based discourse trees. The inherit advantage of generating nuclearity-attributed discourse trees becomes obvious when revisiting the definition of nuclearity in the Rhetorical Structure Theory \cite{marcu1996building, marcu2000theory}, where the nuclearity attribute encodes a notion of ``importance" in the local context, where \textit{Nucleus-Statellite (N-S)} and \textit{Satellite-Nucleus (S-N)} attributions define a directionality between two nodes, while \textit{Nucleus-Nucleus (N-N)} attributions imply equal importance \cite{morey2018dependency}.

Technically, we integrate the nuclearity attribute into the tree-generation process by assigning each subtree one of the three nuclearity classes \textit{N-S}, \textit{S-N} or \textit{N-N}, following the assumption that the attention values $a_{c_{l}}, a_{c_{r}}$ capture the nodes' relative importance in the tree. Starting from the leaf-node attention,
%starts with 
%applying the following strategy:
%We believe that 
%the attention-values
extracted from MILNet, we propagate the attention values through the tree structure according to equation (\ref{eq:functions}).
%we argue that the relative attention of the child nodes is a good approximation for their importance in the local context and can therefore be used as indicators for the nuclearity assignment.
%the attention-values extracted from the MILNet model effectively capture the relative importance of EDUs along with their sentiment. As we already propagate those attention values through the tree structure (see equation \ref{eq:functions}), we argue that the relative attention of the child nodes is a good approximation for their importance in the local context and can therefore be used as indicators for the nuclearity assignment. 
More specifically, for a subtree where the attention value $a_{c_l}$ is greater than the attention $a_{c_r}$, we will assign the \textit{N-S} label, while \textit{S-N} is assigned if the opposite is true. However, in this way, only two of the three possible nuclearity classes can be represented (namely \textit{N-S} and \textit{S-N}), as the attention values are distinct. To further account for the third class of \textit{N-N}, we include an additional subtree at every merge in the CKY procedure, which averages not only the two attention values $a_{c_l}, a_{c_r}$ (as shown in eq. \ref{eq:functions}) %and distributes them uniformly to 
but also the child polarity scores $p_{c_l}, p_{c_r}$. This reflects the definition of the \textit{N-N} nuclearity class according to RST%the Rhetorical Structure Theory
, where an even importance for all child nodes is assumed in the multi-nucleus case.
The additional complexity of doubling the number of trees in each cell is only manageable due to the use of our heuristic approach.

\section{Evaluation}
\label{Evaluation}
In this section, we evaluate our proposed method to generate the MEGA-DT discourse treebank by assessing the performance of a discourse parser when trained on MEGA-DT against our previously proposed ``silver-standard" treebank \cite{huber2019predicting} as well as two commonly used, human-annotated, discourse corpora.
%in regards to the four objectives defined in section \ref{arb_doc}. 
%We will start describing the three datasets used to train and test different methodologies on, followed by an introduction to different discourse parsers we use to train on the generated annotations and compare our performance against. We then shortly introduce the performance metrics used. With the goal to find the best discourse tree generation method among the possible combinations of parameters, we evaluate the different approaches on a subset of the complete dataset and argue for a single set of parameters using empirical evidence on the sample. We then apply the best parameter combination on the complete dataset and show the results of the methodology in a larger context.

\subsection{Treebanks}
\begin{table}
\setlength{\belowcaptionskip}{-10pt}
\centering
\scalebox{.93}{
\begin{tabular}{| l || r | rrr |}
\hline
\multirow{2}{*}{Treebank} & \multirow{2}{*}{\#Documents} & \multicolumn{3}{c|}{\#EDUs*}\\
&& min & avg & max\\
\hline \hline 
Instr-DT & 176 & 2 & 33 & 248 \\
RST-DT & 385 & 2 & 56 & 240 \\
Yelp13-DT & 100,000 & 2 & 10 & 20 \\
MEGA-DT & 250,000 & 2 & 19 & 150 \\
\hline
\end{tabular}}
\caption{Treebank size and distribution (*calculated on the training set)}
\label{tab:datasets}
\end{table}
%We use three publicly available datasets to train and/or test our discourse-tree generation approach. 

The two human-annotated treebanks are: \\
\textbf{Instructional-DT} %This treebank 
(from here on called \textit{Instr-DT}) by \citet{subba2009effective}, which comprises of %encompasses 
documents on home-repair instructions annotated with full RST-style discourse trees, %also %separated 
separated into training- and test-set with a 90-10 split.\\
\textbf{RST-DT} %This treebank 
by \citet{carlson2002rst}, containing %documents in the 
news articles %domain 
alongside 
with full RST-style discourse trees, in the standard 90-10 train-test split.%between training- and test-set.
\medskip

\noindent The two automatically annotated treebanks are: \\
\noindent\textbf{Yelp13-DT}, %We generate Yelp13-DT 
generated according to our previously proposed unconstrained CKY %the non-scalable 
approach as described in \citet{huber2019predicting}. We use the pre-segmented version of the Yelp'13 customer review dataset by \citet{angelidis2018multiple}, separated into EDUs by applying the publicly available discourse segmenter proposed in \citet{feng2014linear}. Yelp13-DT contains short documents with $\leq 20$ EDUs, only considering two nuclearity classes (namely N-S and S-N).\\
%The corpus contains customer-reviews %documents 
%from the 2013 Yelp Dataset Challenge in an 80-10-10 split between training-, development- and test-set. Every document in the corpus contains a single customer-rating on a 5-point star-scale. 
%As the Yelp'13 dataset does not contain any gold-label discourse annotation, we resort to two well-known discourse parsing treebanks to evaluate MEGA-DT. \\
\textbf{MEGA-DT}, our novel treebank, is also generated from the original Yelp'13 corpus, akin to Yelp13-DT. However, due to our newly proposed, scalable solution, MEGA-DT is much larger 
%allowing us to process longer documents 
and more comprehensive, integrating all three nuclearity classes.
%However, MEGA-DT is a much larger and more richly annotated treebank than Yelp-13-DT, because it is built with our novel scalable solution that can process longer documents and also deliver a complete nuclearity annotation.
%To be comparable to the treebank by \citet{huber2019predicting} we generate MEGA-DT similar to Yelp13-DT. %from a pre-segmented version of the Yelp'13 corpus \citep{tang2015document}, published by \citet{angelidis2018multiple}. The corpus contains customer-reviews %documents 
%from the 2013 Yelp Dataset Challenge in an 80-10-10 split between training-, development- and test-set. Every document in the corpus contains a single customer-rating on a 5-point star-scale. 
%As the Yelp'13 dataset does not contain any gold-label discourse annotation, we resort to two well-known discourse parsing treebanks to evaluate Yelp13-DT and MEGA-DT. %\\

\noindent A comparison of the key dimensions of all treebanks used in this work is shown in Table \ref{tab:datasets}. %key dimensions used in this work are summarized in .\\
 
\subsection{Discourse Parsers}
To interpret %show 
our results in the context of existing work, we consider
%compare our system against 
a diverse set of top-performing discourse parsers. Previous work by \citet{morey2017much} compares a set of competitive parsers, including %We evaluate against 
%the \textbf{HILDA} model by \citet{hernault2010hilda}, 
\textbf{DPLP} \citep{ji2014representation}, \textbf{gCRF} \citep{feng2014linear}, \textbf{CODRA} \citep{joty2015codra} and \citet{li2016discourse}. % and \citet{braud2017cross}. 
%the \textbf{DPLP} discourse parser \citep{ji2014representation}, the \textbf{gCRF} model by \citet{feng2014linear}, \textbf{CODRA} by \citet{joty2015codra} as well as the parser by \citet{li2016discourse}. % and \citet{braud2017cross}. 
We further add the \textbf{Two-Stage} discourse parser by \citet{wang2017two} and the neural approach by \citet{yu2018transition} into our final evaluation. Due to the top performance of the parser by \citet{wang2017two} on the structure-prediction of the widely used RST-DT corpus, %of the model by \citet{wang2017two},
and even more importantly, due to the separation of the relation computation from the structure/nuclearity prediction, we use the parser by \citet{wang2017two} in our inter-domain experiments.%new MEGA-DT corpus.
%As a traditional, greedy, bottom-up discourse parser using Support-Vector-Machines (SVMs) the HILDA model by \citet{hernault2010hilda} is still frequently used to compare state-of-the-art models against. We further compare our results against the DPLP discourse parser proposed by \citet{ji2014representation}, using lexical features to build a SVM-based shift-reduce parser, the CODRA approach by \citet{joty2015codra}, combining a CKY chart parser with Dynamic Conditional Random Fields and the state-of-the-art discourse parser by \citet{wang2017two}, using two separate SVMs for structure/nuclearity and relation prediction. We further use the model by \citet{wang2017two} to train on our new large-scale corpus of nuclearity annotated discourse structures.

%\subsection{Metrics}
%To be consistent with previous work in the area, such as \citet{joty2015codra, wang2017two, morey2017much, huber2019predicting}, we report the average micro precision metric for both, the structure and nuclearity measure, in all our experiments.

\subsection{Preliminary Evaluation}
\label{prelim}
%In order to explore different aspects of our %proposed 
%methodology as well as perform hyper-parameter selection,
%parameters of heuristic strategy, tree aggregation, beam-width, algorithms and nuclearity, 
We run a set of preliminary evaluations on a randomly selected subset containing 10,000 documents from the Yelp'13 dataset. 
%The results on this subset are used to obtain a general intuition on the influence of different parameters on the performance. 
%As the true strength of the approach will presumably emerge more clearly when applied to a larger amount of datapoints, 
%These initial %preliminary 
%experiments are only meant to provide initial guidance on the hyper-parameter selection. 
%We omit details for brevity and only show relevant insights here.
In general, the preliminary evaluation suggests that
(1) A beam-size of $10$ delivers the best trade-off between computational complexity and performance (out of \{1, 5, 10, 50, 100\}), when tested according to the distance between gold-label sentiment and model prediction.
(2) 
%To reduce the computational complexity, previous work \citep{huber2019predicting} 
We employ a sentence-first aggregation strategy, using sentence-boundary predictions from the NLTK toolkit\footnote{\url{www.nltk.org/api/nltk.tokenize.html}}. By not allowing inter-sentence connections, unless the complete sentence is already represented by a subtree, we reach superior results in the preliminary evaluation compared to exploring the complete CKY space. This is consistent with previous findings showing that sentence boundaries are key signals for tree aggregations \citep{joty2015codra}.
\subsection{Experiments and Results}
\label{final}
\begin{table*}[ht!]
\setlength{\belowcaptionskip}{-10pt}
\centering
\scalebox{0.86}{
\begin{tabular}{|l|r r | r r|r r | r r|}
\hline
\multirow{3}{*}{Approach} & \multicolumn{4}{c|}{Structure} & \multicolumn{4}{c|}{Nuclearity}\\
 & \multicolumn{2}{c|}{RST-DT} & \multicolumn{2}{c|}{Instr-DT} & \multicolumn{2}{c|}{RST-DT} & \multicolumn{2}{c|}{Instr-DT}\\
 & \multicolumn{1}{c}{Par.} & \multicolumn{1}{c|}{R-Par.} & \multicolumn{1}{c}{Par.} & \multicolumn{1}{c|}{R-Par.} & \multicolumn{1}{c}{Par.} & \multicolumn{1}{c|}{R-Par.} & \multicolumn{1}{c}{Par.} & \multicolumn{1}{c|}{R-Par.}\\
\hline \hline 
Right Branching & 9.27 & 54.64 & 25.45 & 62.72 & $\times$ & $\times$ & $\times$ & $\times$\\
Left Branching & 7.45 & 53.73 & 4.32 & 52.16 & $\times$ & $\times$ & $\times$ & $\times$\\
Hier. Right Branching & \textbf{48.74} & \textbf{74.37} & \textbf{50.68} & \textbf{75.34} & $\times$ & $\times$ & $\times$ & $\times$\\
Hier. Left Branching & 41.16 & 70.58 & 27.50 & 63.75 & $\times$ & $\times$ & $\times$ & $\times$\\
Majority Class & $\times$ & $\times$ & $\times$ & $\times$ & \textbf{\textsuperscript{(N-S)}61.28} & \textbf{\textsuperscript{(N)}61.33} & \textbf{\textsuperscript{(N-N)} 52.33} & \textbf{\textsuperscript{(N)} 76.48} \\
\hline \hline
\multicolumn{9}{|c|}{\textbf{Intra-Domain} Evaluation}\\
\hline
%HILDA\shortcite{morey2017much} & 65.10 & 82.60 & -- & -- & 54.60 & 66.60 & -- & --\\
DPLP\shortcite{ji2014representation}* & 64.10 & 82.00 & -- & -- & 54.20 & 68.20 & -- & --\\
gCRF\shortcite{feng2014linear}* & 68.60 & 84.30 & -- & -- & 55.90 & 69.40 & -- & --\\
CODRA\shortcite{joty2015codra}* & 65.10 & 82.60 & -- & \textbf{82.88}  & 55.50 & 68.30 & -- & \textbf{64.13}\\
%MTL Hier-LSTM\shortcite{morey2017much%braud2016multi} & 59.50 & 79.70 & -- & -- & 47.20 & 63.60 & -- & --\\ 
Li\shortcite{li2016discourse}* & 64.50 & 82.20 & -- & -- & 54.00 & 66.50 & -- & --\\ 
%Braud Cross+Dev\shortcite{braud2017cross}* & 62.70 & 81.30 & -- & -- & 54.50 & 68.10 & -- & --\\ 
Two-Stage\shortcite{wang2017two} & \textbf{70.97} & \textbf{86.00} & \textbf{58.86} & 79.43 & \textbf{57.97} & 72.40 & \textbf{40.00} & 62.39\\
Yu\shortcite{yu2018transition} & -- & 85.50 & -- & -- & -- & \textbf{73.10} & -- & -- \\
\hline \hline
\multicolumn{9}{|c|}{\textbf{Inter-Domain} Evaluation}\\
\hline
Two-Stage\textsubscript{RST-DT} & $\times$ & $\times$ & 45.95 & 73.57 & $\times$ & $\times$ & \textbf{27.18} & 49.78\\
Two-Stage\textsubscript{Instr-DT} & 46.01 & 74.32 & $\times$ & $\times$ & \textbf{22.22} & 44.68 & $\times$ & $\times$\\
Two-Stage\textsubscript{Yelp13-DT\shortcite{huber2019predicting}} & 52.95 & 76.41 & 46.59 & 74.14 & 15.51 & 35.72 & 7.27 & 33.35\\
Two-Stage\textsubscript{MEGA-DT} & \textsuperscript{$\dagger$}\textbf{55.76} & \textsuperscript{$\dagger$}\textbf{77.82} & \textsuperscript{$\dagger$}\textbf{50.23} & \textsuperscript{$\dagger$}\textbf{75.18} & 15.86 & \textbf{44.88} & 20.31 & \textsuperscript{$\dagger$}\textbf{54.87}\\
\hline\hline
Human \shortcite{morey2017much} & 78.70 & 88.30 & -- & -- &  66.80 & 77.30 & -- & --\\
\hline
\end{tabular}}
\caption{Results of the micro-averaged precision measure using the original Parseval method (Par.) %\cite{morey2017much} 
and RST Parseval (R-Par.). %Structure- and nuclearity-predictions are %performance, 
%evaluated on the RST-DT and Instr-DT corpora. 
Inter-domain subscripts identify the training set. Inter-domain results averaged over 10 independent runs. Models with stochastic components are averaged over 3 distinct generation processes. The best performance per sub-table is \textbf{bold}. * Results taken from \citet{morey2017much}, \textsuperscript{$\dagger$} statistically significant with p-value $\leq .05$ to the best inter-domain baseline (Bonferroni adjusted), -- non-published values, $\times$ not feasible combinations}
\label{tab:final}
\end{table*}

%For the final evaluation, 
We train the %established and publicly available SOTA %state-of-the-art 
discourse parser by \citet{wang2017two}\footnote{\url{www.github.com/yizhongw/StageDP/}} %(from here on called \textit{Two-Stage}) 
on our newly generated %discourse-annotated Yelp'13 \citep{tang2015document}  
MEGA-DT corpus as well as the Yelp13-DT and the original RST-DT and Instr-DT corpora\footnote{Trained on an Intel Core i9 (10 Cores, 3.30 GHz) CPU}. To verify the ability of the training treebanks to support the discourse parser in extracting domain-independent features of general discourse, we evaluate the performance on the inter-domain discourse parsing task, training the Two-Stage discourse parser on one domain (e.g., Yelp user reviews in MEGA-DT) and evaluating it on documents in a different domain (e.g., news articles in RST-DT). We compare the obtained performances against the classic and arguably easier intra-domain measure (training and testing on documents within the same domain).
%of the \textit{Two-Stage} parser along with other recent discourse parsers and simple baselines for structure and nuclearity.

The results of the final evaluation are summarized and aggregated in three sets of experiments in Table \ref{tab:final}. In the first set (on top of Table \ref{tab:final}), we show the micro-averaged original Parseval performance (Par.) \cite{morey2017much} as well as the RST-Parseval measures (R-Par.) of %simple 
standard linguistic baselines for the structure- and nuclearity-prediction task. 
Regarding the structure prediction (left), we compare the performance when applying a strictly right- or left-branching tree to the data, as well as hierarchical versions of those (right-/left-branching trees on sentence-level combined by right-/left-branching trees on document level). The results indicate that the hierarchical right-branching tree resembles the original tree structure the closest on both metrics and either evaluation treebank\footnote{Note that the performance of the Hierarchical Right-Branching baseline is higher than %the one
reported in \citet{huber2019predicting}, because of an additional clean-up step required during %the 
data preprocessing. The competitive performance of this baseline is most likely attributed to the highly structured nature of the target domains.
%instructional text.% domain.
}. As a baseline for the nuclearity prediction task, we compute the majority class on the training corpora. %Nucleus-Satellite (61.28\%) for RST and Nucleus-Nucleus (52.33\%) for Instr-DT.
%While for RST-DT the most common nuclearity-class  is Nucleus-Satellite (61.28\%), it is Nucleus-Nucleus (52.33\%) on Instr-DT. %Please note that while we use the majority class as a simple baseline in this work, it cannot be directly compared with the output of a discourse-parser, as it requires additional information (supervision signals) on the target domain, which are mostly unavailable.
It is important to note that while the linguistic baselines for structure do not require available training data, the %one for nuclearity, being the 
majority class measure depends on access to an annotated corpus in the target domain. %Therefore, it is fair to directly compare the performance of the majority baseline only with the results of intra-domain experiments and not the inter-domain evaluation, where the parser %by design 
%has no access to annotated data in the target domain.
 
\begin{table*}[t!]
\setlength{\belowcaptionskip}{-10pt}
\centering
\scalebox{0.86}{
\begin{tabular}{|l|r r | r r|r r | r r|}
\hline
\multirow{3}{*}{Approach} & \multicolumn{4}{c|}{Structure} & \multicolumn{4}{c|}{Nuclearity}\\
 & \multicolumn{2}{c|}{RST-DT} & \multicolumn{2}{c|}{Instr-DT} & \multicolumn{2}{c|}{RST-DT} & \multicolumn{2}{c|}{Instr-DT}\\
 & \multicolumn{1}{c}{Par.} & \multicolumn{1}{c|}{R-Par.} & \multicolumn{1}{c}{Par.} & \multicolumn{1}{c|}{R-Par.} & \multicolumn{1}{c}{Par.} & \multicolumn{1}{c|}{R-Par.} & \multicolumn{1}{c}{Par.} & \multicolumn{1}{c|}{R-Par.}\\
\hline \hline 
Two-Stage\textsubscript{MEGA-DT-Base} & 51.85  & 75.92 & 47.05 & 73.87 & 17.22 & 36.46 & 8.64 & 34.48\\
Two-Stage\textsubscript{MEGA-DT +Stoch} & 55.05 & 77.58 & 43.75 & 73.76 & \textbf{17.76} & 37.43 & 8.22 & 35.89\\
Two-Stage\textsubscript{MEGA-DT +Nuc} & 54.55 & 76.76 & 50.01 & 74.35 & 13.82 & 44.22 & 19.82 & 54.10\\
Two-Stage\textsubscript{MEGA-DT} & \textbf{55.76} & \textbf{77.82} & \textbf{50.23} & \textbf{75.18} & 15.86 & \textbf{44.88} & \textbf{20.31} & \textbf{54.87}\\
\hline 
\end{tabular}}
\caption{Ablation study showing the influence of nuclearity and stochasticity on the overall performance, measured as the micro-average precision using original Parseval (Par.) and RST Parseval (R-Par.). Results averaged over 10 runs (using 3 distinct generation processes if a stochastic components is included). The best performance is \textbf{bold}.}
\label{tab:ablation}
\end{table*}
 
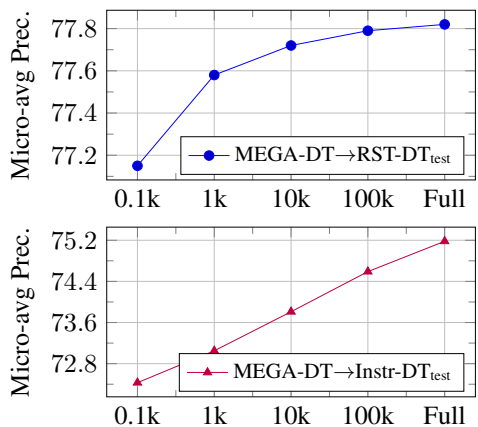
\begin{figure}
\setlength{\belowcaptionskip}{-16pt}
  \centering
  \scalebox{0.9}{
  \begin{tikzpicture}
  \begin{axis}[
    width=0.7*\linewidth, 
    height=2.5cm, 
    scale only axis,
    ylabel={Micro-avg Prec.},
    at={(0.015\linewidth,0)},
    anchor=above north west,
    ytick distance=0.2,
    minor tick num=1,
    xtick=data,
    xticklabels={0.1k, 1k, 10k, 100k, Full},
    legend pos=south east,
    legend style={font=\small},
    ymajorgrids=true,
    xmajorgrids=true,
]
    \addplot
        coordinates {
        	(0,77.15)
        	(1,77.58)
        	(2,77.72)
        	(3,77.79)
        	(4,77.82)
        	};
    \addlegendentry{MEGA-DT$\rightarrow$RST-DT\textsubscript{test}}
    
    \end{axis}    
\end{tikzpicture}}
\label{fig:rst_results}
  \centering
  \scalebox{0.9}{
  \begin{tikzpicture}
  \begin{axis}[
    width=0.7*\linewidth, 
    height=2.5cm, 
    scale only axis,
    ylabel={Micro-avg Prec.},
    at={(0.015\linewidth,0)},
    anchor=above north west,
    ytick distance=.8,
    minor tick num=1,
    xtick=data,
    xticklabels={0.1k, 1k, 10k, 100k, Full},
    legend pos=south east,
    legend style={font=\small},
    ymajorgrids=true,
    xmajorgrids=true,
]
    \addplot[mark=triangle*,color=purple]
        coordinates{
        	(0,72.43)
        	(1,73.05)
        	(2,73.81)
        	(3,74.59)
        	(4,75.18)
        	};
    \addlegendentry{MEGA-DT$\rightarrow$Instr-DT\textsubscript{test}}
    
    \end{axis}    
\end{tikzpicture}}
%\begin{tikzpicture}
%\begin{axis}[
%    ybar,
%	bar width=4pt,
%	height=4cm,
%    width=0.95*\linewidth, 
%    ylabel={Variance},
%    xlabel={Dataset Size},
%    legend style={font=\small},
%    yticklabel=\pgfmathparse{\tick}\pgfmathprintnumber{\pgfmathresult}\,\%, %   
%    xtick=data,
%    ymax=5,
%    ytick distance=1,
%    xticklabels={0.1k, 1k, 10k, 100k, Full},
%    ymajorgrids=true,
%    xmajorgrids=true,
%    ]
%    \addplot+[ybar] plot 
%    coordinates {
%        (0,2.38)
%        (1,2.689)
%        (2,2.668)
%        (3,1.953)
%        (4, 1.573)
%    };
%    \addlegendentry{MEGA-DT$\rightarrow$RST-DT\textsubscript{test}}
%
%    \addplot+[ybar] plot 
%    coordinates {
%        (0,2.477)
%        (1,3.613)
%        (2,2.136)
%        (3,1.795)
%        (4, 1.249)
%    };
%    \addlegendentry{MEGA-DT$\rightarrow$Instr-DT\textsubscript{test}}
%    
%\end{axis}
%\end{tikzpicture}
\caption{Performance-trend over increasingly large subsets tested on RST-DT (top) and Instr-DT (bottom). Each sample is generated as the average performance over 10 random subsets, drawn from 3 independently created treebanks.}
\label{fig:size_analysis}
\end{figure}
 
The second set of results shows the intra-domain performance of top performing discourse parsers, frequently evaluated against in the past. %. The results thereby show the 
%in predicting %on the intra-domain task for
%both structure  (%on the 
%left) and  nuclearity  (%on the 
%right). 
While all parsers except CODRA \citep{joty2015codra} have been only evaluated on RST-DT, we additionally train and evaluate the Two-Stage parser on the Instr-DT corpus. When comparing the intra-domain discourse parsing performance, the Two-Stage parser reaches the consistently best performance on RST-DT structure prediction, while the discourse parser by \citet{yu2018transition} achieves the best results on the RST-DT nuclearity prediction using RST-Parseval. CODRA reaches the best performance on the Instr-DT corpus when evaluated with RST-Parseval.%\footnote{The original Parseval scores on the Instr-DT dataset cannot be compared due to missing results from previous parsers.}

The main contribution of this work is placed in the third set of results, where the Two-Stage discourse parser is trained and tested on different, non-overlapping domains (i.e., inter-domain).
This task is arguably more useful and significantly more difficult than the task evaluated in the second set, which is reflected in the performance decrease for structure and nuclearity in the first two rows of the sub-table, confirming that the transfer of discourse-structures and -nuclearity between domains is a challenging task. The results presented in the third row of the sub-table show the performance of the Two-Stage parser when trained on Yelp13-DT, containing short documents with limited nuclearity annotations. %recent model by \citet{huber2019predicting}, trained on a $100,000$ datapoint subset of the Yelp'13 corpus, limited to short documents. 
The approach achieves consistently better performance compared to the first two rows on the inter-domain structure prediction task (For both, original Parseval and RST-Parseval), as %claimed in the original paper.
we have previously shown in \citet{huber2019predicting}. However, only considering two out of three nuclearity classes (N-S and S-N), the system performs rather poorly on the nuclearity classification task.
The bottom row of the third sub-table displays the %remarkable
performance of the Two-Stage discourse parser when trained on our new MEGA-DT corpus. %which was
%generated using the scalable stochastic beam-search in combination with the novel nuclearity attribution. %is evaluated in the bottom row of the third sub-table. 
Training on MEGA-DT %Our novel system 
delivers statistically significant improvements %better performance
over the best inter-domain baseline 
in all structure prediction tasks. Furthermore, our new system also achieves statistically significant gains on the Instr-DT nuclearity prediction, when evaluated according to the RST-Parseval metric. The nuclearity measure on RST-DT using RST-Parseval is %, where training on MEGA-DT %produces is nevertheless 
statistically equivalent %insignificant but on average better results than
to %the system 
the best baseline system.
%training on the human annotated Instr-DT treebank. 
%We compute the significance within a significance interval of $\leq .05$ across 10 independent runs of the discourse parser, using three different stochastic initializations for models containing the stochastic component.
%with $77.64\%/74.5\%$ on the structure- and $44.74\%/54.29\%$ on the nuclearity-prediction on RST-DT and Intr-DT respectively.
%While the quantitative analysis presented in Table \ref{tab:final} strongly suggests the superiority of our MEGA-DT corpus over human annotated discourse treebanks when applied on the challenging inter-domain discourse-parsing task, 
Overall, our MEGA-DT corpus %generated using the stochastic approximation and the added nuclearity attribute shows the largest improvement over 
appears to outperform previously published treebanks for inter-domain discourse parsing on every sub-task on at least one competitive metric.

In order to gain deeper insights into the effectiveness of our proposed treebank generation approach, we run a set of four additional evaluations.
First, we evaluate the individual components of our system by showing an ablation study in Table \ref{tab:ablation}, starting with the performance of the discourse parser trained with MEGA-DT-Base, a treebank generated %without relying on the stochastic component
with the standard beam-search approach and without integrating nuclearity. Adding each feature separately (+Stoch, +Nuc) we observe improvements on at least one of the sub-tasks; however, the combination of the two components produces the best performing MEGA-DT corpus.
Second, we show the performance-trend over increasingly large subsets of MEGA-DT in Figure \ref{fig:size_analysis}, tested on RST-DT (top) and Instr-DT (bottom). The two trends highlight consistent improvements with increasingly large dataset sizes, suggesting further possible gains with even larger treebanks.
Third, we further analyze the nuclearity classification performance in Table \ref{tab:confusion}, which presents four confusion matrices for the discourse parsing output of our MEGA-DT treebank, evaluated according to the original Parseval and RST-Parseval metrics on RST-DT and Instr-DT. The matrices show a potential explanation for the performance-gap between the original Parseval and the RST-Parseval metrics, identifying the over-prediction of the N-N class, especially for gold-label N-S nuclearities. Further, we frequently misclassify the gold-label N-S nuclearity class as S-N.
%While we cannot reach the majority class baseline on RST-DT, the results when compared to models trained on human-annotated datasets show competitive results...}
Lastly, we present an additional qualitative analysis in Appendix \ref{appendix_1} %selecting a random subset of %1,000  discourse trees from our MEGA-DT corpus. 
to investigate the strength and potential weaknesses of trees in MEGA-DT. %the automatically generated treebank, 
We therefore show three randomly selected trees that closely/poorly reflect the authors gold-label sentiment respectively (see Figure \ref{fig:teaser} for a teaser). In general, the qualitative analysis shows that trees in MEGA-DT are non-trivial, reasonably balanced, strongly linked to the EDU-level sentiment and mostly well-aligned with meaningful discourse-structures.

\begin{table}
\setlength{\belowcaptionskip}{-10pt}
    \begin{minipage}{.5\linewidth}
      \centering
      \scalebox{0.8}{
        \setlength{\extrarowheight}{5pt}
        \begin{tabular}{cc|c|c|c|}
          & \multicolumn{1}{c}{} & \multicolumn{3}{c}{Predicted} \\
          & \multicolumn{1}{c}{} & \multicolumn{1}{c}{N-N}  & \multicolumn{1}{c}{N-S}  & \multicolumn{1}{c}{S-N} \\\cline{3-5}
          %\multirow{3}{*}{ \rotatebox{90}{Gold (\%)}} & N-N & 18.9 & 0.9 & 3.1 \\ \cline{3-5}
          \multirow{3}{*}{ \rotatebox{90}{Gold}} & N-N & 243 & 11 & 40 \\ \cline{3-5}
          %& N-S & 44.4 & 2.2 & 7.5 \\\cline{3-5}
          & N-S & 570 & 28 & 96 \\\cline{3-5}
          %& S-N & 19 & 0.4 & 3.7 \\\cline{3-5}
          & S-N & 244 & 5 & 47 \\\cline{3-5}
        \end{tabular}}
    \end{minipage}%
    \begin{minipage}{.5\linewidth}
      \centering
      \scalebox{0.8}{
       \setlength{\extrarowheight}{5pt}
        \begin{tabular}{cc|c|c|}
          & \multicolumn{1}{c}{} & \multicolumn{2}{c}{Predicted} \\
          & \multicolumn{1}{c}{} & \multicolumn{1}{c}{N}  & \multicolumn{1}{c}{S} \\ \cline{3-4}
              %& N & 54.2 & 5.2 \\ \cline{3-4}
              & N & 1947 & 186 \\ \cline{3-4}
              %& S & 36.8 & 3.9 \\ \cline{3-4}
              & S & 1321 & 141 \\ \cline{3-4}
    \end{tabular}}
\end{minipage} 
\begin{minipage}{.5\linewidth}
      \centering
      \scalebox{0.8}{
    \setlength{\extrarowheight}{5pt}
    \begin{tabular}{cc|c|c|c|}
      & \multicolumn{1}{c}{} & \multicolumn{1}{c}{N-N}  & \multicolumn{1}{c}{N-S}  & \multicolumn{1}{c}{S-N} \\\cline{3-5}
         %\multirow{3}{*}{ \rotatebox{90}{Gold (\%)}}   & N-N & 39.2 & 0.5 & 4.1 \\ \cline{3-5}
         \multirow{3}{*}{ \rotatebox{90}{Gold}}   & N-N & 85 & 1 & 9 \\ \cline{3-5}
         %& N-S & 35.5 & 1 & 7.4 \\\cline{3-5}
         & N-S & 77 & 2 & 16 \\\cline{3-5}
         %& S-N & 10.6 & 0.5 & 1.4 \\\cline{3-5}
         & S-N & 23 & 1 & 3 \\\cline{3-5}
    \end{tabular}}
\end{minipage}%
    \begin{minipage}{.5\linewidth}
      \centering
      \scalebox{0.8}{
       \setlength{\extrarowheight}{5pt}
        \begin{tabular}{cc|c|c|}
          & \multicolumn{1}{c}{} & \multicolumn{1}{c}{N}  & \multicolumn{1}{c}{S} \\ \cline{3-4}
              %& N & 71.6 & 4.6 \\ \cline{3-4}
              & N & 470 & 30 \\ \cline{3-4}
              %& S & 22.7 & 1.1 \\ \cline{3-4}
              & S & 149 & 7 \\ \cline{3-4}
    \end{tabular}}
\end{minipage} 
\caption{Confusion Matrices for the model trained on MEGA-DT, evaluated on RST-DT (top) and Instr-DT (bottom). Left: Original Parseval, Right: RST-Parseval}
\label{tab:confusion}
\end{table}

\section{Conclusions and Future Work}
\label{Conclusions and Future Work}
In this work, we present a novel distant supervision approach to predict the discourse-structure and -nuclearity for documents of arbitrary length solely using document-level sentiment information. To deal with the increasing spatial complexity, we apply and compare heuristic beam-search strategies, %motivated by recent through previous applications of heuristic approximations on tree-structured natural language data
including a stochastic variant inspired by RL techniques. %, through the strong similarities between the task-objective and reward sparsity in RL and the task at hand. 
Our results on the challenging inter-domain discourse-structure and -nuclearity prediction task strongly suggests that the heuristic approach taken
%taken in this paper %\textcolor{red}{
(1) enhances the structure prediction task through more diversity in the early-stage tree selection, (2) allows us to effectively predict nuclearity and (3) helps to significantly reduce the complexity of the unrestricted CKY approach to scale for arbitrary length documents.
%}

In conclusion, %With the superior performance shown, we give evidence that 
our new approach allows the NLP community to augment any existing sentiment-annotated dataset with discourse trees, enabling the automated generation of large-scale domain/genre-specific discourse treebanks. As a case study for the effectiveness of the approach, we annotate and publish our MEGA-DT corpus as a high quality RST-style discourse treebank, which has been shown to outperform previously 
%proposed large-scale dicourse corpora and small-scale human 
proposed discourse treebanks (namely Yelp13-DT, RST-DT and Instr-DT) on most tasks of inter-domain discourse parsing. This suggests that parsers trained on our MEGA-DT corpus (or further domain-specific treebanks generated according to our approach) should be used to
%for the task of 
derive discourse trees in target domains where no gold-labeled data is available.
%without reaching the computational limitations and impairing the performance.

This work can be extended in several ways: (i) We plan to investigate into further functions for $\tau$ to enhance the exploration-exploitation trade-off. (ii) Additional strategies to assign nuclearity should be explored, considering the excessive N-N-classification shown in our evaluation. (iii) We plan to apply our approach to more sentiment datasets (e.g., \citet{diao2014jointly%, zhang2015character
}), creating even larger treebanks. (iv) Our new and scalable solution can be extended to also predict discourse relations besides structure and nuclearity. (v) We also plan to use a neural discourse parser (e.g. \citet{yu2018transition}) in combination with our large-scale treebank to fully leverage the potential of data-driven discourse parsing approaches. (vi) Taking advantage of the new MEGA-DT corpus, we want to revisit the potential of discourse-guided sentiment analysis, to enhance current systems, especially for long documents%with rich discourse information
. (vii) Finally, more long term, 
we intend to explore other auxiliary tasks for distant supervision of discourse, like summarization, question answering and machine translation, for which plenty of annotated data exists (e.g., \citet{RNN_beyond, cohan-etal-2018-discourse, rajpurkar2016squad, rajpurkar2018know}).

\begin{figure}[h!]
\setlength{\belowcaptionskip}{-10pt}
    \centering
    \includegraphics
    [width=.82\linewidth]
    {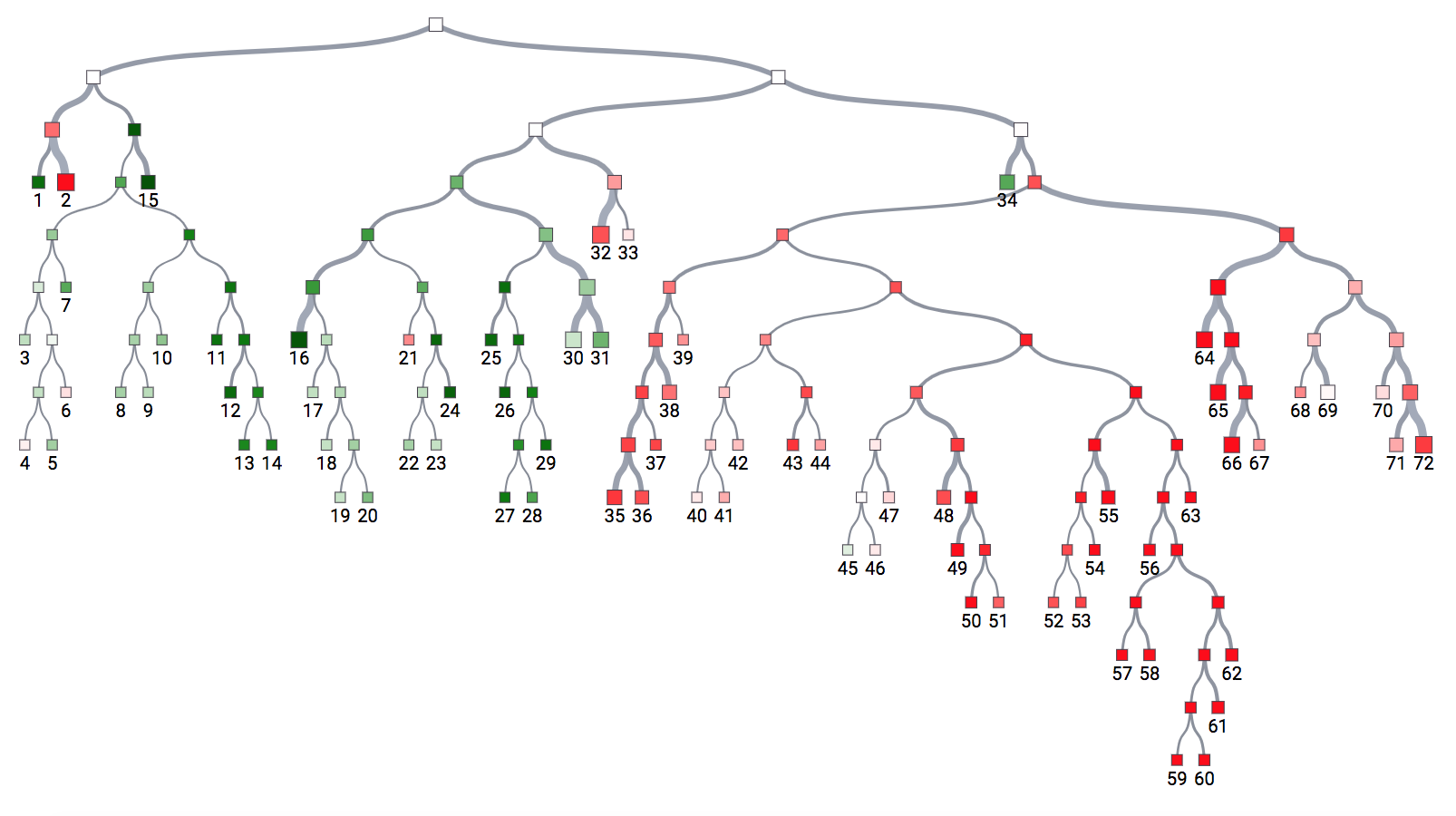}
    \caption{Teaser for a tree analyzed in Appendix \ref{appendix_1} containing 72 EDUs and neutral document-level sentiment.}
    \label{fig:teaser}
\end{figure}

\section*{Acknowledgments}
We thank %the anonymous 
reviewers and the UBC-NLP group for their insightful comments. % and suggestions. 
This research was supported by the
Language \& Speech Innovation Lab of Cloud BU, Huawei Technologies Co., Ltd.

\bibliographystyle{acl_natbib}
\bibliography{emnlp2020}

\clearpage

\appendix
%\onecolumn
\section{Qualitative Analysis of Generated Discourse Trees}
\label{appendix_1}
The following examples are automatically generated trees from our MEGA-DT corpus. EDU leaf-nodes are enumerated and can be referenced with the discourse units in the description. The colour-saturation and -hue values represents the sentiment of the nodes, with a dark red (high saturation) representing a strongly negative subtree, white (low saturation) representing a neutral sentiment subtree and a dark green (high saturation) represents a strongly positive subtree. The thickness of edges and the size of nodes represent the attention of the subtree, which is strongly correlated with the subtree nuclearity.

\begin{figure*}[h!]
    \centering
    \includegraphics[width=1\linewidth]{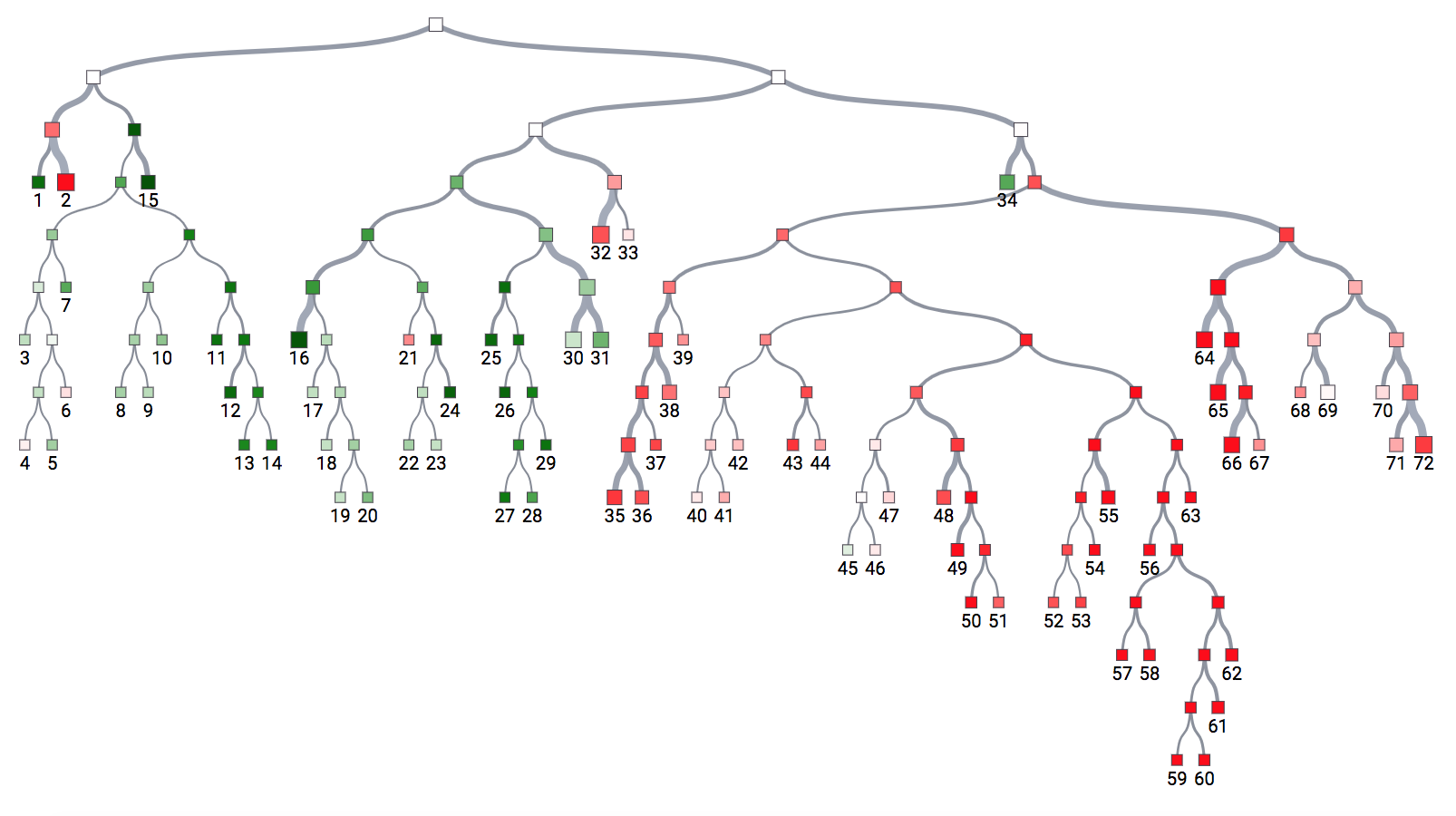}
    \caption{Accurately predicted example, Gold-label polarity: $0$, Predicted polarity: $0.03$, \\
    Discourse: [amazing food.]$_{1}$, [awful, awful service.]$_{2}$, [the garlic bread. very good.]$_{3}$, [softer than i expected,]$_{4}$, [which was nice.]$_{5}$, [i also just wasn't expecting garlic bread.]$_{6}$, [so it was a nice surprise.]$_{7}$, [escargot -]$_{8}$, [i was the only one at the table (of 10)]$_{9}$, [to eat it.]$_{10}$, [they were great!]$_{11}$, [served bubbling hot, not rubbery at all, delicious sauce.]$_{12}$, [i kept the dish]$_{13}$, [to dip bread into just because of the sauce.]$_{14}$, [veal - amazing.]$_{15}$, [everything tasted fantastic.]$_{16}$, [ok, the carrots]$_{17}$, [that were on the side were a bit plain]$_{18}$, [and could have been softer, but the veal itself and the sauce]$_{19}$, [it was in, and the mushrooms and pasta.]$_{20}$, [i left nothing on my plate.]$_{21}$, [my husband got the same]$_{22}$, [and also had the same impression.]$_{23}$, [creme brulee - fantastic.]$_{24}$, [tasted great, good texture.]$_{25}$, [pleasantly surprised.]$_{26}$, [my husband got the tiramisu]$_{27}$, [and said]$_{28}$, [it was great.]$_{29}$, [so why the 3 stars]$_{30}$, [when the food was so amazing?]$_{31}$, [because of the terrible service. 1 -]$_{32}$, [we got water.]$_{33}$, [great.]$_{34}$, [but our server * never * asked us]$_{35}$, [if we wanted anything else.]$_{36}$, [when my husband finally stopped him to ask for a glass for my father in law, a coke for]$_{37}$, [and other drinks, our server looked very inconvenienced by it. 2 -]$_{38}$, [didn't get to order appetizers.]$_{39}$, [you see]$_{40}$, [i got escargot?]$_{41}$, [i ordered that with my meal.]$_{42}$, [our server never asked about appetizers]$_{43}$, [and went straight to meals.]$_{44}$, [also, my husband was walking with our daughter]$_{45}$, [when the ordering was starting]$_{46}$, [and needed an extra minute.]$_{47}$, [our server wanted to start with him.]$_{48}$, [when asked if he could start with someone else's order,]$_{49}$, [our server protested,]$_{50}$, [but eventually did move on to the next person.]$_{51}$, [you'd think]$_{52}$, [starting at the next person was]$_{53}$, [asking him to cut off his hand. 3 - empty glasses everywhere!]$_{54}$, [never got or was offered a refill on my drink.]$_{55}$, [or anyone else's.]$_{56}$, [when my father stopped our server well]$_{57}$, [after our meal was over]$_{58}$, [and asked]$_{59}$, [if i could get a coke,]$_{60}$, [our server said]$_{61}$, [i had never ordered one.]$_{62}$, [well of course i hadn't.]$_{63}$, [i never had a chance to! 4 -]$_{64}$, [offering dessert seemed a complete afterthought.]$_{65}$, [will i recommend this place to anyone else?]$_{66}$, [conditionally.]$_{67}$, [i'll make sure to tell them]$_{68}$, [that the food was very good, but not to go]$_{69}$, [if they want attentive service,]$_{70}$, [are on any kind of time constraint, expect refills on their drinks,]$_{71}$, [or are at all shy about getting a server's attention.]$_{72}$}
\end{figure*}

\begin{figure*}%101028
    \centering
    \includegraphics[width=.8\linewidth]{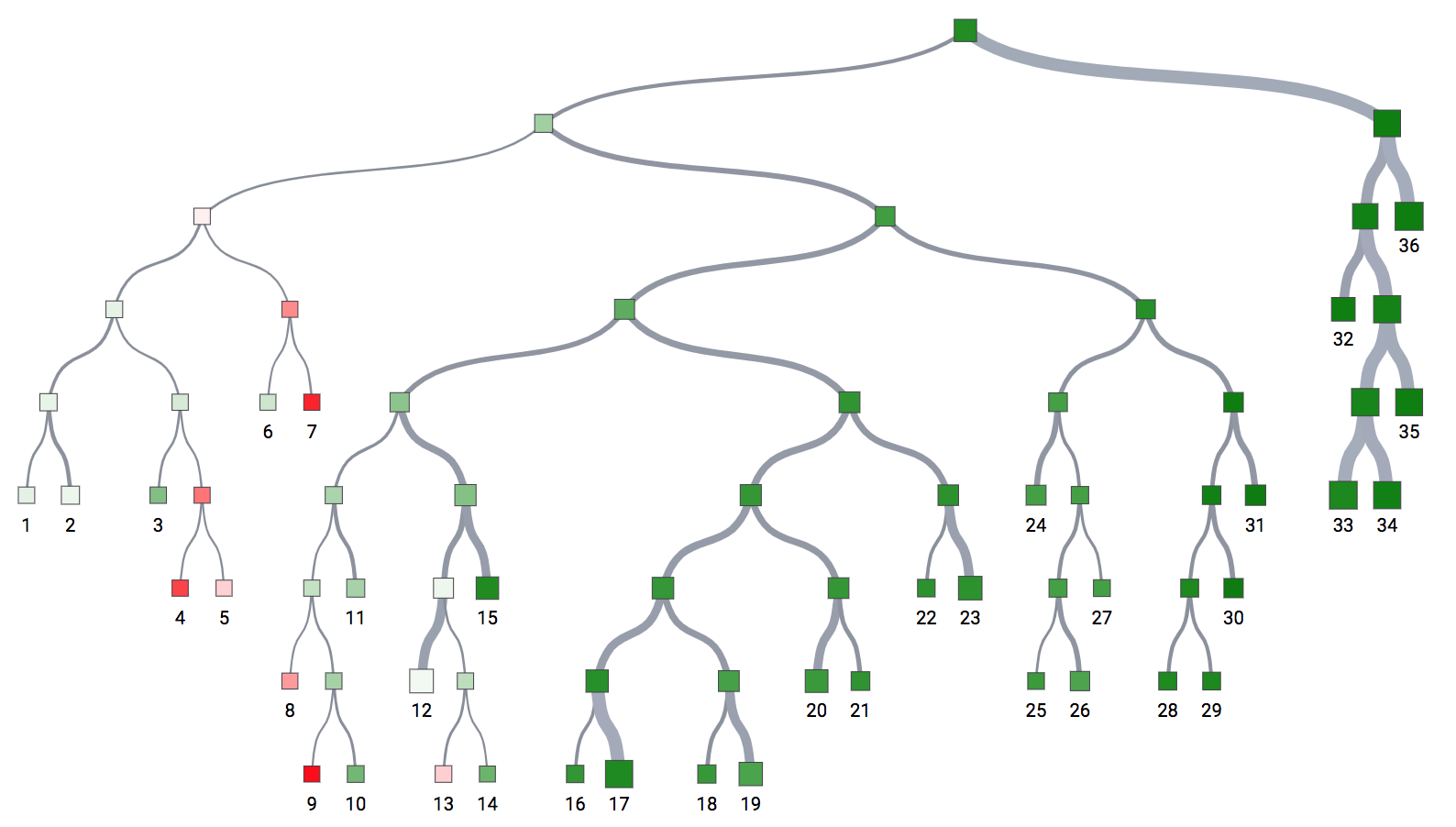}
    \caption{Accurately predicted example, Gold-label polarity: $0.5$, Predicted polarity: $0.345$, \\
    Discourse: [it has been a hell of a work week]$_{1}$, [and friday could not have come any sooner.]$_{2}$, [this week was rough especially with the announcement]$_{3}$, [that we are officially in a recession]$_{4}$, [and the ambiguity was hitting me from all sides.]$_{5}$, [man, did]$_{6}$, [i need some distraction from all my worries.]$_{7}$, [so the bf ( bacon ) and i decided on thai]$_{8}$, [but wanted to venture out from the norm]$_{9}$, [and we are very glad]$_{10}$, [we did.]$_{11}$, [the thai hut exceeded our expectations.]$_{12}$, [we were a little skeptical at first with so many lackluster reviews]$_{13}$, [but we hit the jackpot on our night.]$_{14}$, [this place has a great vibe !]$_{15}$, [we were seated immediately]$_{16}$, [and staff was beyond courteous and attentive.]$_{17}$, [we were approached by a few staff members]$_{18}$, [which gave us the feeling of true teamwork.]$_{19}$, [our server was attentive]$_{20}$, [and even sparked up some conversation throughout our meal.]$_{21}$, [we started with a hot pot of chicken tom kha kai]$_{22}$, [and this soup hit the spot.]$_{23}$, [my worries were vanishing with every spoonful.]$_{24}$, [they say]$_{25}$, [chicken soup will cure a cold]$_{26}$, [and that menudo will feed a hangover.]$_{27}$, [well, i]$_{28}$, [now believe]$_{29}$, [tom kha kai is the cure for the blues]$_{30}$, [because it sure made me happy !]$_{31}$, [for our main dish we shared the red chicken curry.]$_{32}$, [a little heavy on the red and green peppers but very tasty and was the perfect match with the soup]$_{33}$, [so we will definitely be back]$_{34}$, [and will be sharing this place with some of our closest friends.]$_{35}$, [i've already made lunch plans for next wednesday.]$_{36}$}
    \label{fig:my_label}
\end{figure*}

\begin{figure*}%100038
    \centering
    \includegraphics[width=.6\linewidth]{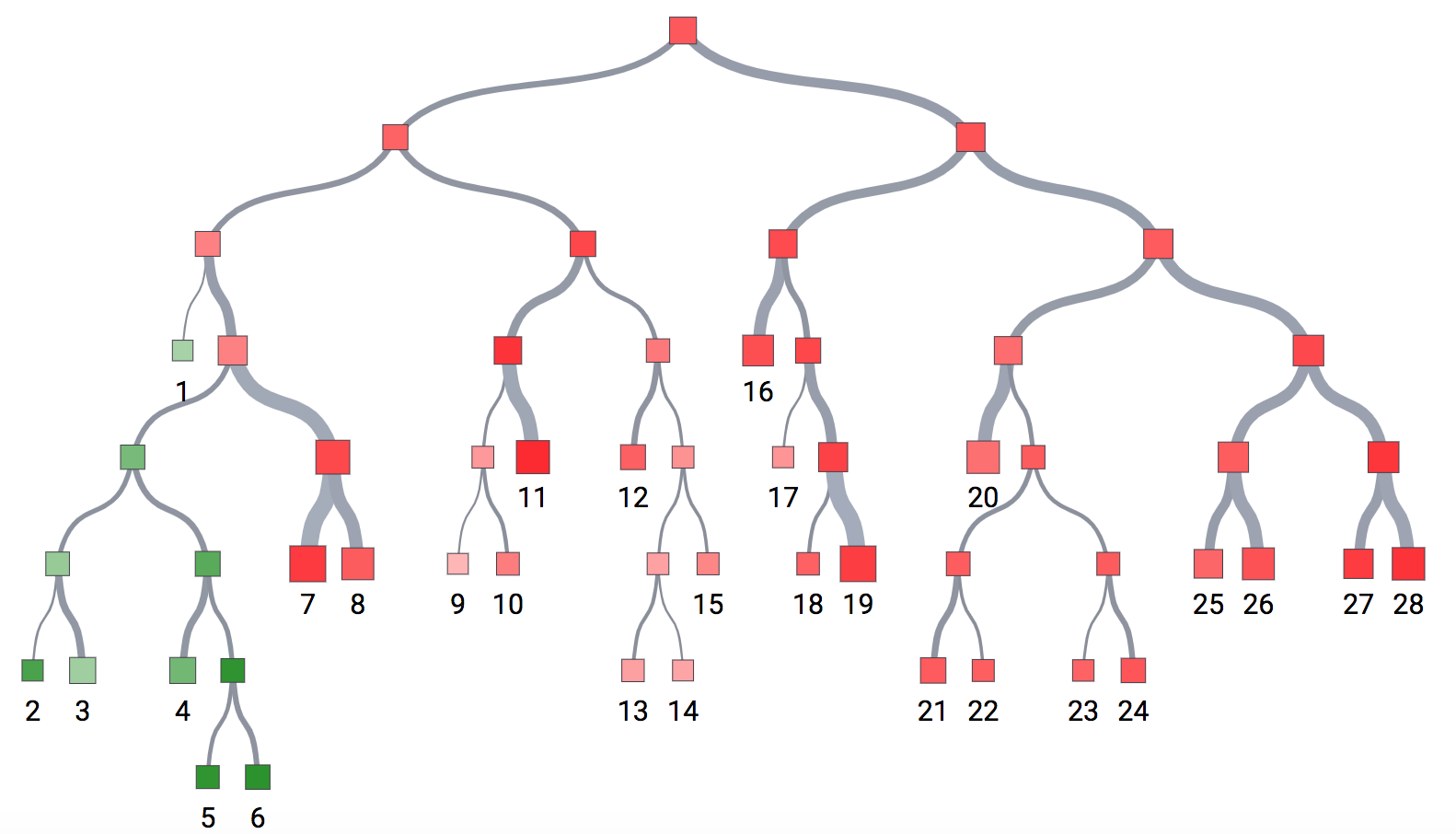}
    \caption{Accurately predicted example, Gold-label polarity: $-0.5$, Predicted polarity: $-0.391$, \\
    Discourse: [stopped in here for a friday happy hour with co-workers.]$_{1}$, [the beer was decently]$_{2}$, [priced for happy hour.]$_{3}$, [the appetizers were decently priced,]$_{4}$, [which would be awesome]$_{5}$, [if they were good.]$_{6}$, [the chicken strips were terrible.]$_{7}$, [i have never eaten something so greasy and yet dry all at once.]$_{8}$, [they are beer battered ( like fish )]$_{9}$, [which could be good,]$_{10}$, [but the execution on this was terrible.]$_{11}$, [the outside was really greasy]$_{12}$, [which took away all of the crispy goodness]$_{13}$, [that usually happens]$_{14}$, [when things are battered and deep fried.]$_{15}$, [the chicken itself was dry as a bone.]$_{16}$, [we also got an order of fries]$_{17}$, [that came out cold]$_{18}$, [and were just below mediocre.]$_{19}$, [the place was really warm,]$_{20}$, [which could be attributed to the summer heat,]$_{21}$, [but we were sitting inside,]$_{22}$, [so there is a fair assumption]$_{23}$, [that air conditioning would be involved.]$_{24}$, [i'll pass next time]$_{25}$, [my coworkers are planning a trip here.]$_{26}$, [i'd be better off]$_{27}$, [eating at mcdonald's.]$_{28}$}
\end{figure*}

\begin{figure*}%105242
    \centering
    \includegraphics[width=.65\linewidth]{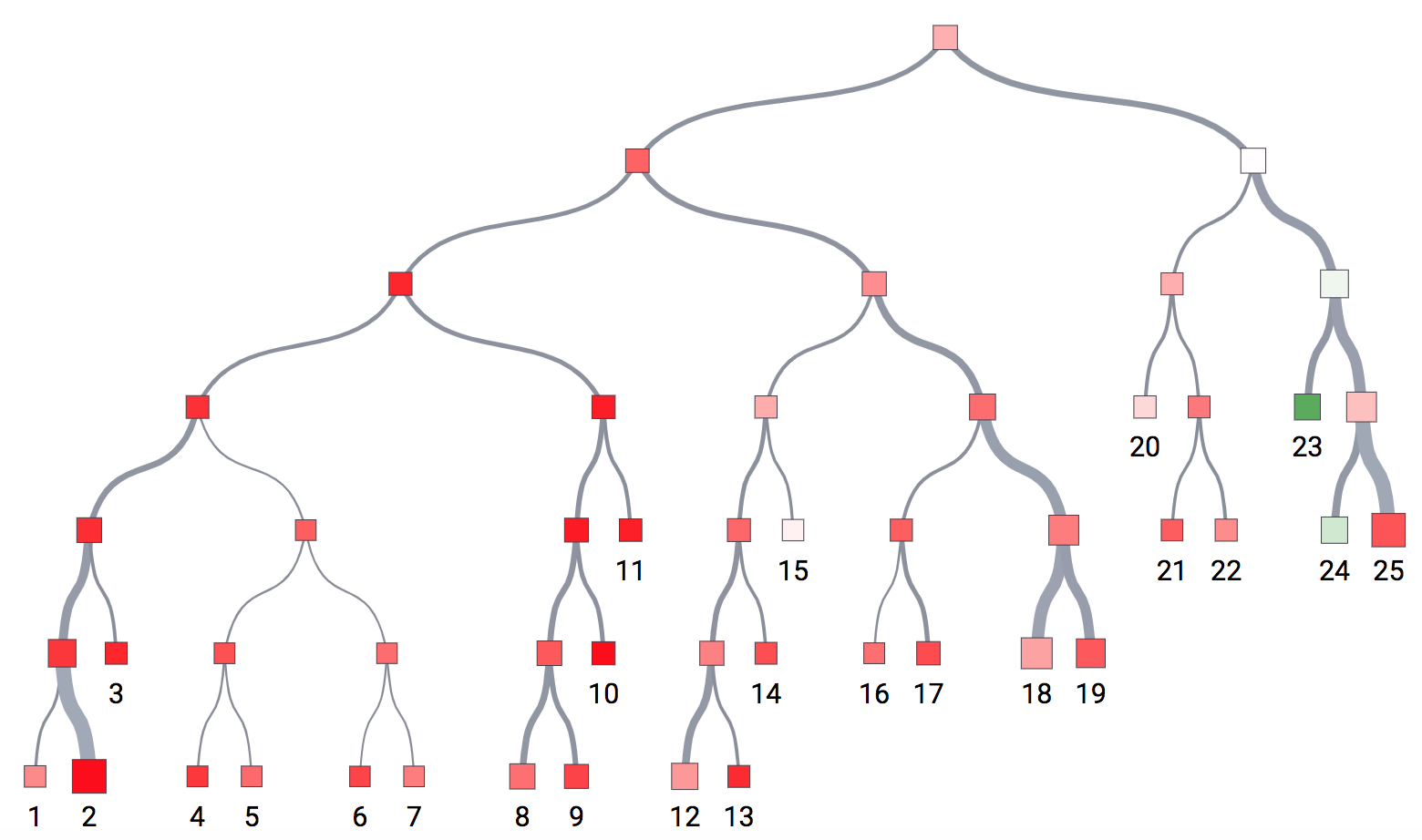}
    \caption{Inaccurately predicted example, Gold-label polarity: $1$, Predicted polarity: $-0.098$, \\
    Discourse: [upon first moving here 2 years ago,]$_{1}$, [i had the worse experience]$_{2}$, [attempting to get an airbrush spray tan at this salon.]$_{3}$, [they had only 2 people at specific times]$_{4}$, [that could spray you custom.]$_{5}$, [no problem showed up]$_{6}$, [and the tech could not figure out how to use the gun.]$_{7}$, [so awkward enough]$_{8}$, [him being a male]$_{9}$, [and standing there naked, i to get my money back]$_{10}$, [after waiting 20 min.]$_{11}$, [well a couple months back]$_{12}$, [they ran a deal for versa,]$_{13}$, [which is a booth spray tan.]$_{14}$, [i love this booth.]$_{15}$, [it is like airbrushing but private,]$_{16}$, [and this spray tan absolutely does not smell or stain your sheets!]$_{17}$, [i found this]$_{18}$, [upon leaving denver, co. and just]$_{19}$, [until i saw it online on living deals for amazon. :)]$_{20}$, [one down two]$_{21}$, [to go.]$_{22}$, [all for \$ 29 :) love !]$_{23}$, [as far as the gym goes,]$_{24}$, [never used it !]$_{25}$}
\end{figure*}

\begin{figure*}%10053
    \centering
    \includegraphics[width=.65\linewidth]{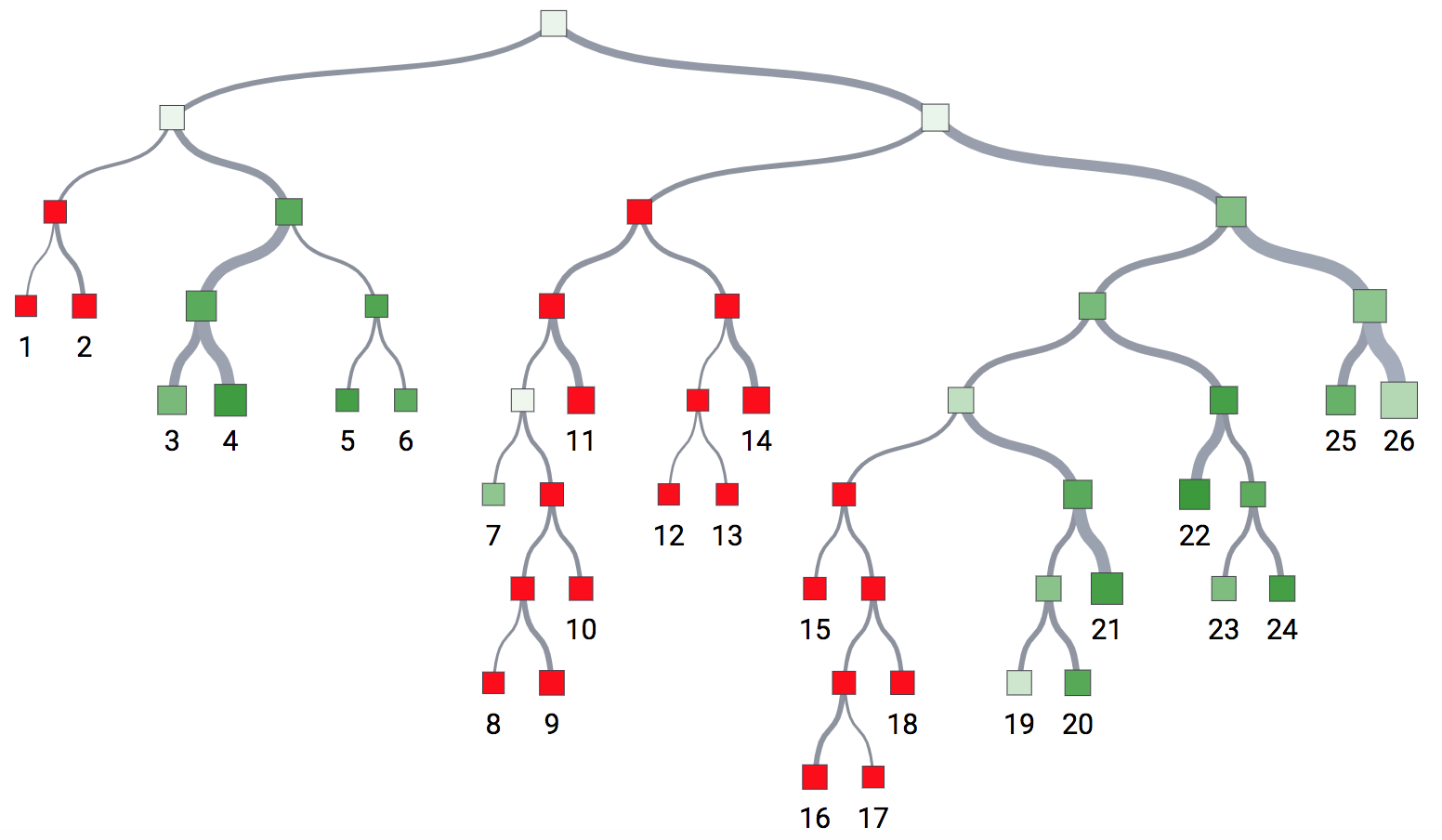}
    \caption{Inaccurately predicted example, Gold-label polarity: $-0.5$, Predicted polarity: $0.085$, \\
    Discourse: [i hate having]$_{1}$, [to write a poor review for this joint!]$_{2}$, [the owner is a really great guy]$_{3}$, [and the service was excellent.]$_{4}$, [the place is decorated well]$_{5}$, [and has a clean finished look.]$_{6}$, [i really wanted to love the pudding]$_{7}$, [but it really did n't work out for my wife and i. from first glance]$_{8}$, [the pudding was all very soupy]$_{9}$, [and while it tasted]$_{10}$, [okay, was not anything to write home about.]$_{11}$, [the shop is trying too hard]$_{12}$, [to be an ice cream or gelato setup.]$_{13}$, [i think all the flavors and take away from their core business model.]$_{14}$, [i think]$_{15}$, [they should focus on making the rice pudding more solid]$_{16}$, [and have a couple]$_{17}$, [warm pudding options.]$_{18}$, [i can envision a warm rice pudding with some nuts and raisins with some brown sugar or cinnamon on top]$_{19}$, [yum !]$_{20}$, [shoot for rich, creamy and full of flavor.]$_{21}$, [my best wishes go out to them]$_{22}$, [and hope]$_{23}$, [that the masses will enjoy it more than we did.]$_{24}$, [they are good folks]$_{25}$, [and deserve to be successful.]$_{26}$}
\end{figure*}

\begin{figure*}
    \centering
    \includegraphics[width=.85\linewidth]{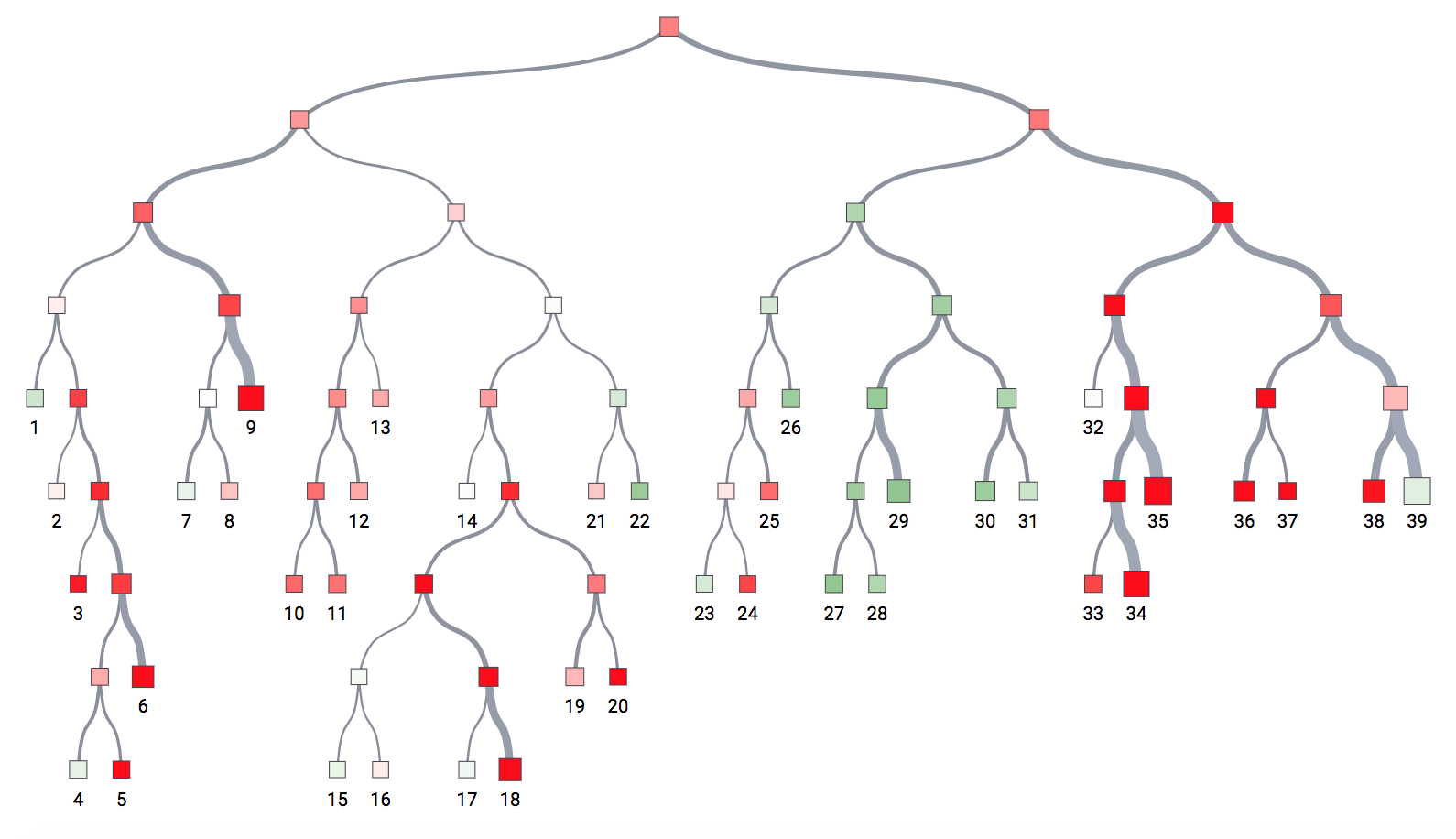}
    \caption{Inaccurately predicted example, Gold-label polarity: $-0.5$, Predicted polarity: $-0.006$, \\
    Discourse: [i've been here a couple of times in the past.]$_{1}$, [usually at someone else's suggestion.]$_{2}$, [i ca n't say]$_{3}$, [that i recommend this place,]$_{4}$, [unless you like]$_{5}$, [your lunch served up with a lot of attitude.]$_{6}$, [the lady]$_{7}$, [that takes the orders at the counter]$_{8}$, [is usually abrasive and rude.]$_{9}$, [i am the type of person]$_{10}$, [who will kill the meanest person with kindness,]$_{11}$, [but there are places]$_{12}$, [where i draw the line.]$_{13}$, [so, i have drawn the line with rome's pizza.]$_{14}$, [the funny part about it all is]$_{15}$, [that my line is often zig-zag and curvy,]$_{16}$, [so i still go here]$_{17}$, [when someone else wants to go.]$_{18}$, [hehe.]$_{19}$, [my friends like the abuse i guess.]$_{20}$, [one friend says]$_{21}$, [the lady is nice to him.]$_{22}$, [the plus side]$_{23}$, [of going here is the fact]$_{24}$, [that they serve an average pizza by the slice with your custom toppings.]$_{25}$, [they also make hoagies and some other dishes.]$_{26}$, [they have a nice lunch special]$_{27}$, [that includes soda for a few bucks.]$_{28}$, [they also serve some typical american favorites like hot wings.]$_{29}$, [i usually order the of pizza lunch special]$_{30}$, [and get the unsweetened tea.]$_{31}$, [i'm not sure]$_{32}$, [why i forget,]$_{33}$, [but their tea tastes horrible]$_{34}$, [because the water from the fountain tastes terrible !]$_{35}$, [but it never fails, i forget]$_{36}$, [that i need to of water with me.]$_{37}$, [all in all, this place is a dive.]$_{38}$, [give it a try.]$_{39}$}
\end{figure*}

\end{document}